\documentclass[10pt,twocolumn,letterpaper]{article}

\usepackage{titling}
\usepackage{iccv}
\usepackage{times}
\usepackage{epsfig}
\usepackage{graphicx}
\usepackage{amsmath}
\usepackage{amssymb}
\usepackage{caption}
\usepackage[accsupp]{axessibility} 
\usepackage{booktabs}
\usepackage{multirow}

\newcommand{\norm}[1]{\left\|#1\right\|}

\usepackage[breaklinks=true,bookmarks=false]{hyperref}

\iccvfinalcopy 

\def\iccvPaperID5474 
\def\httilde{\mbox{\tt\raisebox{-.5ex}{\symbol{126}}}}

\ificcvfinal\fi

\begin{document}

\title{Novel-view Synthesis and Pose Estimation for Hand-Object Interaction\\ from Sparse Views}

\author{
Wentian Qu$^{1,2}$\qquad 
Zhaopeng Cui$^{3}$\qquad
Yinda Zhang$^{4}$\qquad
Chenyu Meng$^{1,2}$\qquad
Cuixia Ma$^{1,2}$\\
Xiaoming Deng$^{1,2}$$\thanks{indicates corresponding author.}$\qquad
Hongan Wang$^{1,2}$$^*$\\
$^1$Institute of Software, Chinese Academy of Sciences \quad $^2$University of Chinese Academy of Sciences\\ 
$^3$State Key Lab of CAD$\&$CG, Zhejiang University  \quad $^4$Google
}

\maketitle
\ificcvfinal\thispagestyle{empty}\fi

\begin{figure*}[ht]
\centering%
\includegraphics[width=\linewidth]{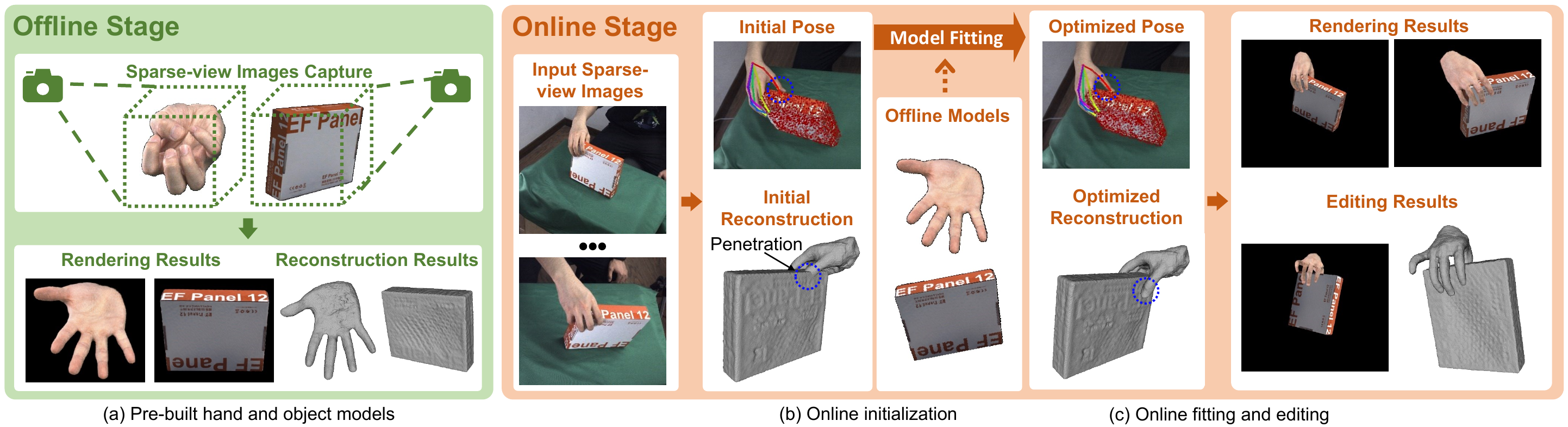}
\captionof{figure}{We propose a neural rendering and pose estimation system for hand-object interaction using sparse view images. (a) During offline stage, we learn hand and object models that enable rendering and shape reconstruction. During  online stage, we initialize the pose from sparse camera views (b), and then conduct online fitting to improve pose estimation, which enables photo-realistic free viewpoint rendering (c). Our framework also naturally supports hand object interaction editing.
}
\label{fig:teaser}
\end{figure*}

\begin{abstract}
Hand-object interaction understanding and the barely addressed novel view synthesis are highly desired in the immersive communication, whereas it is challenging due to the high deformation of hand and heavy occlusions between hand and object. In this paper, we propose a neural rendering and pose estimation system for hand-object interaction from sparse views, which can also enable 3D hand-object interaction editing.
We share the inspiration from recent scene understanding work that shows a scene specific model built beforehand can significantly improve and unblock vision tasks especially when inputs are sparse, and extend it to the dynamic hand-object interaction scenario and propose to solve the problem in two stages.
We first learn the shape and appearance prior knowledge of hands and objects separately with the neural representation at the offline stage. During the online stage, we design a rendering-based joint model fitting framework to understand the dynamic hand-object interaction with the pre-built hand and object models as well as interaction priors, which thereby overcomes penetration and separation issues between hand and object and also enables novel view synthesis. In order to get stable contact during the hand-object interaction process in a sequence, we propose a stable contact loss to make the contact region to be consistent. 
Experiments demonstrate that our method outperforms the state-of-the-art methods. Code and dataset are available in project webpage \href{https://iscas3dv.github.io/HO-NeRF}{https://iscas3dv.github.io/HO-NeRF}.
\end{abstract}

\section{Introduction}
\label{sec:intro}
Hand-object interaction understanding plays an important role in immersive contextual teaching applications such as surgical operation and training in the use of machinery. Previous works mostly focus on the hand-object interaction detection~\cite{FirstPersonAction_CVPR2018}, reasoning~\cite{kwon2021h2o} or pose estimation~\cite{hasson2019learning,hasson2020leveraging}.
However, the barely addressed novel view synthesis of hand-object interaction is also highly desired.

Recently, neural rendering is emerging to facilitate the novel view synthesis simply by learning from a collection of images and produces promising high-quality images. 
Although existing neural rendering approaches perform well on static scenes~\cite{Nerf,mipnerf}, rigid objects~\cite{Object_Nerf,guo2020objectcentric} and human models~\cite{Neural_Body,Animatable_NeRF,a-nerf}, they barely considered scene context in interaction (such as contact~\cite{zhang2020place} and model penetration~\cite{cao2021reconstructing,Grasping_Field}).
In the realm of hand, LISA~\cite{lisa} is the only hand neural rendering model, and achieves promising rendering results of bare hands. However, LISA cannot work well for hand-object interaction due to heavy inter-occlusions and it requires dense (about 20) camera views that may refrain it from wide applications.
It is even more challenging to use sparse-view images to synthesize novel views~\cite{ibrnet,infonerf} and estimate accurate pose for hand-object interaction, which plays a key role in many applications such as Holoportation~\cite{orts2016holoportation} and manipulation skill learning from human demonstration~\cite{qin2022one,arunachalam2022dexterous}.

In this work, we propose a novel-view synthesis and pose estimation system for hand-object interaction scenes with sparse camera views (Fig.~\ref{fig:teaser}). Recent scene understanding work~\cite{nerf_in_room} shows a scene specific model built  beforehand can significantly improve and unblock vision tasks especially when inputs are sparse, and we extend it from static objects to dynamic hand-object interaction scenes and solve the problem in two stages.
We first use sparse-view images as input to train the pose-driven neural rendering models of hand and object during the offline stage.
Benefiting from the progress of hand pose tracking \cite{han2022umetrack,learnable-triangulation,mediapipe} and object pose estimation \cite{cosypose}, we only need very low cost to build hand model and object model. 
Then at the online stage, we estimate both hand and object poses using a novel differentiable rendering-based model fitting under geometric constraints. In this way, we can understand hand-object interaction accurately and render novel views effectively.

However, it is non-trivial to fulfill this goal. 
Firstly, it is difficult to build neural rendering systems from sparse camera views due to insufficient visual information and 
depth ambiguity caused by hand-object inter-occlusions. 
Existing few-shot neural rendering methods \cite{infonerf,ibrnet} fail under sparse camera views (Fig.~\ref{fig:render_quality}). In order to solve this problem, we establish the fitting process based on the pre-built models which provides strong shape and appearance priors, and it can achieve excellent novel view rendering from sparse views. 
Secondly, it is difficult to obtain accurate hand and object poses and reasonable interactions only by photometric constraints due to extensive occlusion. 
In order to handle this problem, we propose a novel differentiable rendering-based model fitting process under geometric constraints to refine the poses and enforce the spatial context between hand and object by leveraging the signed distance function (SDF) to reduce penetration and to encourage tight hand-object regions to contact. We propose a stable contact loss to penalize large sliding of the hand-object contact area across temporally consecutive frames. Through the joint model fitting process, we can achieve accurate pose estimation for hand-object interactions.
Thirdly, our system needs a dataset to include images of hand, object, and hand-object interaction. However, existing datasets such as
\cite{chao2021dexycb,hampali2020honnotate} cannot satisfy this requirement, so we need to collect a real dataset to evaluate our method.

\vspace{-0.5mm}
Our main contribution is summarized as follows. First, to the best of our knowledge, 
we present the first solution to unblock hand-object interaction neural rendering from sparse views. We design a new two-stage approach (i.e. offline model building and online model fitting) to achieve accurate hand-object pose estimation  and photo-realistic novel view synthesis. 
Second, we leverage effective geometric constraints to conduct rendering-based model fitting, which can recover reasonable hand-object interaction even under sparse views and inter-occlusions. To reduce the sliding that occurs during the interaction, we design a new stable contact loss to enforce the hand-object contacted regions to be consistent for video sequences.
Third, we propose a hand-object interaction dataset \emph{HandObject} for neural rendering tasks, including images for hand, object and hand-object interaction scenes.
Finally, experiments demonstrate that, with the help of offline and online stages, our method achieves significantly better performance in pose estimation and rendering quality than previous methods.

\section{Related Work}
\label{sec:relatedwork}
\noindent \textbf{Hand-Object Interaction Understanding.}
Existing hand-object interaction understanding  works~\cite{cao2021reconstructing,jiang2021hand,yang2021cpf,zhang2021manipnet,zhang2021single,hasson2020leveraging,kwon2021h2o,hampali2020honnotate,hasson2019learning,Grasping_Field,a1,a2,a3,a4,a5,a6} usually use hand statistical shape model such as MANO~\cite{mano}, and adopt known object shape to estimate object 6D pose. The geometric feature extracted from implicit network can also be used to represent object shape~\cite{a1,a3}.
The key challenges of hand-object understanding include occlusion, penetration and separation between hand and object. 
However, mesh-based hand and object representation is expensive for surface-based penetration and contact loss~\cite{hasson2019learning}.
In order to deal with the contact for hand-object interaction understanding, several existing works~\cite{hasson2019learning,yang2021cpf,jiang2021hand,zhang2021single,Grasping_Field} achieve reasonable contact by predicting hand regions where contact is likely to occur 
and optimizing the distance between the object vertices and contact regions on the hand.
Recently, implicit shape representations such as SDF~\cite{Grasping_Field,cao2021reconstructing,a1,a5} are emerged to facilitate the detection of the penetration and contact between hand and object, because SDF can indicate the spatial relationship between point and surface and the penetration of two shapes can be judged with the sign of SDF values. Our method adopts SDF representation to facilitate geometric constraints in hand-object interaction.

\begin{figure*}[ht]
\centering%
\includegraphics[width=0.98\linewidth]{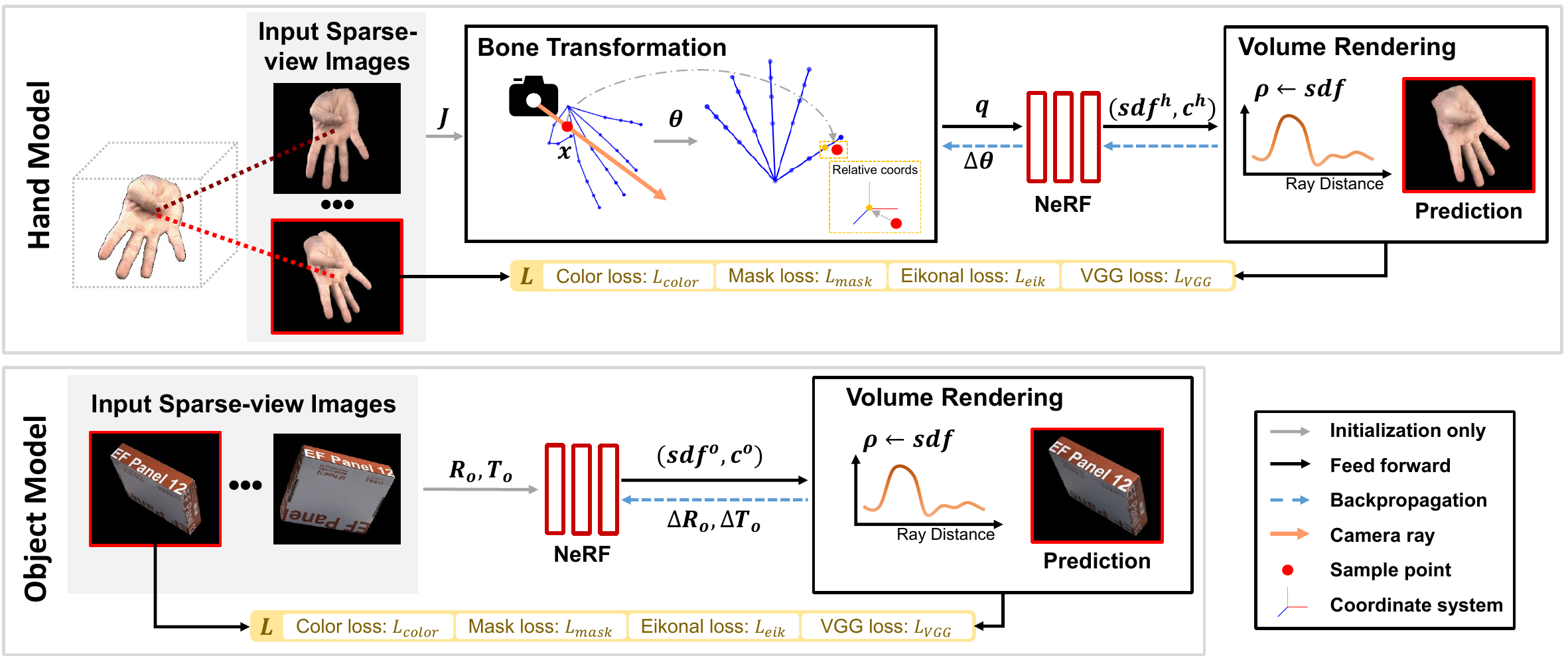}
\caption{Offline stage to learn hand and object models. Top: Hand model. Each sampling point on the ray is converted to the local coordinate system of each hand part through bone transformation. Then we encode the point into embedding vectors and feed it to hand model to get the SDF and color value. Bottom: Object model. We convert the sampling point to the model coordinate with the object pose, and get the SDF and color value. 
}
\label{fig:pipeline}
\end{figure*}
\noindent \textbf{Neural Rendering.}
Neural radiance fields (NeRF)~\cite{Nerf} aims to synthesize novel view of a scene via volume rendering with densely sampled posed images.
The few-shot novel view synthesis methods~\cite{pixelNerf,ibrnet,infonerf,dietnerf} use sparse view images as input to build neural radiation fields, but fail in widely divergent camera view.
Although the shape surface of an object is implicitly included in NeRF, the traditional density cannot extract accurate surface~\cite{neus}. Therefore, IDR~\cite{IDR}, NeuS~\cite{neus} and VolSDF~\cite{volsdf} use SDF or occupancy fields combined with volume rendering to achieve high precision reconstruction. But these methods only reconstruct rigid objects.
From the perspective of new scene synthesis, several NeRF editing works ~\cite{pumarola2021dnerf,Nerfies,Object_Nerf,GIRAFFE} has been proposed. However, these methods cannot be trivially extended to edit non-rigid objects such as human hand.  

\noindent \textbf{Neural Articulated Shape Representation.}
Prior arts~\cite{NASA,Leap,SNARF,skeleton_hand,Neural_Body,Animatable_NeRF,NARF} use implicitly shape representation to reconstruct articulated human body shape.
NASA~\cite{NASA} and its variants~\cite{Leap,SNARF} generates human body shape by converting the posed human body to the canonical pose and querying the SDF value of a point in the canonical pose space. 
In order to learn generative novel view synthesis, several methods~\cite{Neural_Body,Animatable_NeRF,NARF,a-nerf, humannerf_google,structured_liu,arah,lisa} integrate bone transformation~\cite{a-nerf} and linear blend skinning~\cite{Animatable_NeRF,structured_liu,arah,lisa} with neural radiance fields. 
Different from the neural rendering systems~\cite{Neural_Body,NARF,a-nerf}, we present 
a hand-object interaction neural rendering method using sparse-view images, in which spatial context between hand and object are modeled by the SDF representations to reduce penetration and encourage stable contact.
Compared to implicit articulated shape representation such as NASA~\cite{NASA}, Animatable NeRF~\cite{Animatable_NeRF} and DD-NeRF~\cite{ddnerf} that need parametric shape models or skinning weight supervision, our method can learn the geometry and appearance of hand and object with sparse-view only.

\section{Method}
\label{sec:method}
\begin{figure*}[t]
\centering%
\includegraphics[width=0.98\linewidth]{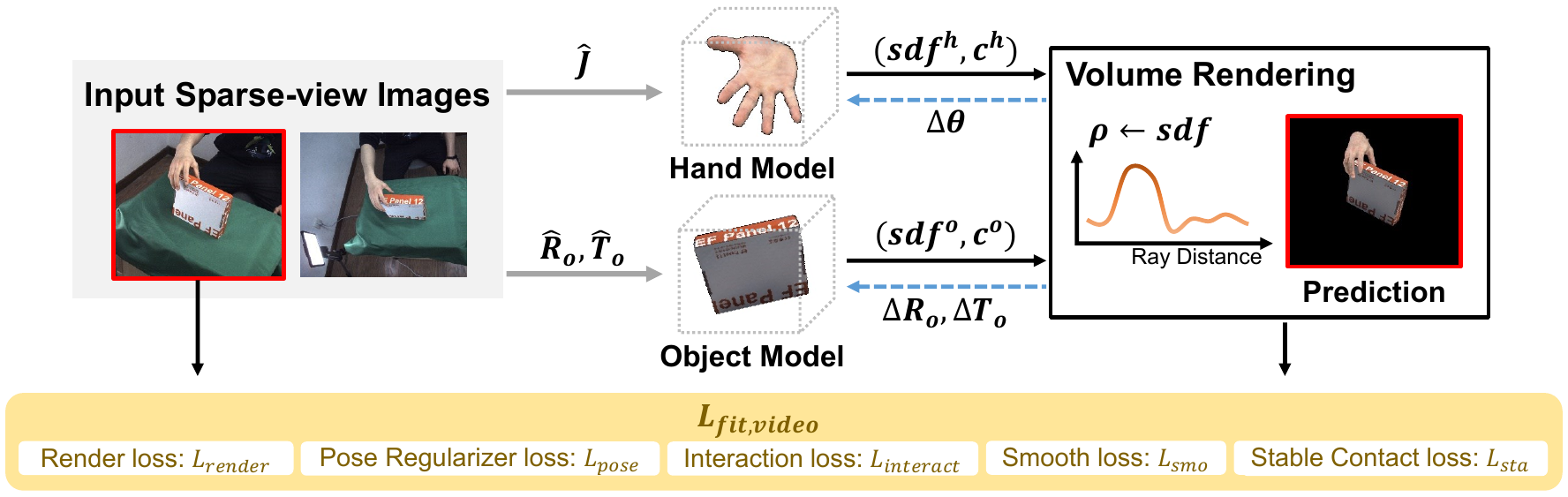}
\caption{Online stage for joint model fitting. We utilize hand/object pose estimation networks for initialization, and refine pose with $\mathcal{L}_{fit}$ for single frame and $\mathcal{L}_{fit,video}$ for video sequence. 
}
\label{fig:fitting}
\end{figure*}

Given sparse-view observations of hand-object interaction,
we aim to generate free-viewpoint synthesis of the scene and estimate hand skeleton pose and object 6D pose.
Our framework is divided into two stages: offline model building and online model fitting.
At the offline stage, we learn the neural models for hand and object individually based on the pre-captured sparse-view images (Sec.~\ref{sec:3-2}).
As shown in Fig.~\ref{fig:pipeline}, our neural hand model is a generative implicit representation driven by hand skeleton pose, which can be used to represent geometry and to generate novel views and our object model can be driven by object 6D pose. 
At the online stage (Sec.~\ref{sec:3-3}), given the sparse-view images, we estimate both hand and object poses using a rendering-based model fitting under effective geometric constraints (Fig.~\ref{fig:fitting}).
For video input, we can further enforce smooth and stable contact loss to reduce pose jitters between frames to generate more smooth and consistent hand-object interaction.
Benefited from our offline models and online fitting method, we can also edit the hand-object interaction scenes (Sec.~\ref{sec:4-3}).

\subsection{Offline Stage for Hand-Object Model Building}
\label{sec:3-2}
\vspace{1mm}
\noindent \textbf{Hand Neural Rendering Model.}
We aim to build a pose-driven hand model that can achieve novel view synthesis and recover accurate geometry (Top of Fig.~\ref{fig:pipeline}).
In our hand model, we convert sampling point on camera ray to local coordinate system of each hand part through bone transformation. Then we encode the point into embedding vectors and feed it to hand model to get the SDF and color value.
We use SDF-based implicit fields for hand geometry representation~\cite{neus}, because it enables more accurate surface than NeRF's density fields and facilitates geometric constraints during model fitting (Sec.~\ref{sec:3-3}).
Our hand model can be formulated as:
\begin{equation}
\begin{aligned}
    &f_{hand}^{s}(\mathbf{x}(t),\mathbf{J})=sdf^{h},F^{h},\\
    &f_{hand}^{c}(\mathbf{x}(t), \mathbf{J},F^{h},\mathbf{n}_{h})=\mathbf{c}^{h},
    \label{eq:hand_model}
\end{aligned}
\end{equation}
where $\mathbf{x}(t)=\mathbf{o}+t\mathbf{d}$ represents the sampling point on the ray, $\mathbf{o}\in\mathbb{R}^{3}$ is the camera optical center, $\mathbf{d}\in\mathbb{R}^{3}$ is the ray direction, $sdf^{h} \in \mathbb{R}$ represents SDF value under hand model, $\mathbf{n}_{hand} \in \mathbb{R}^{3}$ represents the derivation of $sdf^{h}$, $\mathbf{c}^{h} \in \mathbb{R}^{3}$ represents the color value, $f$ represents the MLP network and $F^{h}$ represents the features output by $f_{hand}^{s}$.
We define the hand skeleton pose as $\mathbf{J} \in \mathbb{R}^{n_{j}\times3}$ ($n_{j}$=$21$ is the hand joint number), which can be used to calculate the bone transformation $\mathbf{B}^{-1} \in \mathbb{R}^{n_{j}\times4\times4}$ and the position of each joint in canonical pose $\mathbf{T} \in \mathbb{R}^{n_{j}\times3}$ as used in HALO~\cite{skeleton_hand}. 
Thus each sampling point $\mathbf{x}$ can be converted to the local bone coordinate systems by:
\begin{equation}
\begin{bmatrix}
    \mathbf{q}\\
    1
\end{bmatrix}
= \mathbf{B}^{-1}
\begin{bmatrix}
    \mathbf{x}\\
    1
\end{bmatrix}
-
\begin{bmatrix}
    \mathbf{T}\\
    1
\end{bmatrix}.
\end{equation}

Inspired by A-NeRF~\cite{a-nerf}, we calculate the components of the embedding vectors by  $\mathbf{v}=\norm{\mathbf{q}}_{2}$, $h(\mathbf{v})=1-S(\tau(\mathbf{v}-\bar{\mathbf{v}}))$ and $\mathbf{r}=\frac{\mathbf{q}}{\mathbf{v}}$, where $S(\cdot)$ represents the Sigmoid operation, $\bar{\mathbf{v}}$ represents the cutoff point, $\tau$ represents the sharpness, and $\gamma(\cdot)$ represents the positional encoding.
We combine them as embedding vectors $\mathbf{e}^{s}_{h}=[h(v)\gamma(v),h(v)\gamma(\mathbf{r})]$ and feed it into $f^{s}_{hand}$ to get SDF value. Then we add $\mathbf{n}_{hand}$ in the embedding vector as $\mathbf{e}^{c}_{h}=[h(v)\gamma(v),h(v)\gamma(\mathbf{r}), F^{h}, \gamma(\mathbf{n}_{h})]$ and feed it into $f^{c}_{hand}$ to get color value. Similar to NeuS~\cite{neus}, we use $sdf^{h}$ to get opaque density $\rho^{h}$ by:
\begin{equation}
\rho^{h}(t) = \max\left(\frac{-\frac{{\rm d}\Phi_z}{{\rm d} t}(f^{s}_{hand}(\mathbf{x}(t)))}{\Phi_z(f^{s}_{hand}(\mathbf{x}(t)))}, 0\right),
  \label{eq:density}
\end{equation}
where $\Phi_{z}(x)=(1+e^{-zx})^{-1}$ and $z$ is a learnable scalar. More details can be found in the supplementary material.

\vspace{1mm}
\noindent \textbf{Object Neural Rendering Model.}
Our goal is to generate an object model that can be controlled by the object 6D pose (Bottom of Fig.~\ref{fig:pipeline}). 
We can use object 6D pose $\mathbf{R}_{o}\in \mathbb{R}^{3\times3}$ and $\mathbf{T}_{o}\in\mathbb{R}^{3}$ to transform the object from the object coordinate to the world coordinate. 
For each sampling point $\mathbf{x}$, we first convert it to the object model coordinate system with inverse transformation $\mathbf{q}_{o}=\mathbf{R}_{o}^{-1}(\mathbf{x}-\mathbf{T}_{o})$, then encode $\mathbf{q}_{o}$ to the embedding vector, and feed it to the object model to get SDF $sdf^{o}$ and color $\mathbf{c}^{o}$ values.
Our object model can be formulated as:
\begin{equation}
\begin{aligned}
    &f^{s}_{obj}(\mathbf{x}(t), \mathbf{R}_{o},\mathbf{T}_{o})=sdf^{o},F^{o},\\ &f^{c}_{obj}(\mathbf{x}(t), \mathbf{R}_{o},\mathbf{T}_{o},F^{o},\mathbf{n}_{o})=\mathbf{c}^{o}.
    \label{eq:obj_model}
\end{aligned}
\end{equation}

The ray direction under object coordinate system can be expressed by $\mathbf{l}_{o}=\mathbf{R}_{o}^{-1}\mathbf{d}$, and we define the normal of $\mathbf{x}$ in object model as $\mathbf{n}_{o}$. The embedding feature vector of $f^{s}_{obj}$ can be formulated as: $e^{s}_{o}=[\gamma(\mathbf{q}_{o})]$ and the embedding feature vector of $f^{c}_{obj}$ can be formulated as: $e^{c}_{o}=[\gamma(\mathbf{q}_{o}),\gamma(\mathbf{l}_{o}), F^{o},\gamma(\mathbf{n}_{o})]$. 

\vspace{1mm}
\noindent \textbf{Loss Function in Offline Stage.}
To learn hand and object models, we mainly use color loss and mask loss to encourage rendered color $\hat C$ and mask $\hat M$ to be closed to the ground truth respectively.
We also use Eikonal loss~\cite{gropp2020implicit} to regularize the SDF and follow~\cite{keypointnerf} to use VGG loss.
Therefore, the total loss of hand model can be formulated as:
\begin{equation}
  \mathcal{L}=\lambda_{co}\mathcal{L}_{color}+\lambda_{m} \mathcal{L}_{mask}+\lambda_{e} \mathcal{L}_{eik} + \lambda_{v} \mathcal{L}_{VGG},
  \label{eq17}
\end{equation}
where $\mathcal{L}_{color}$,  $\mathcal{L}_{mask}$ and $\mathcal{L}_{eik}$ are similar to NeuS~\cite{neus}, 
$\lambda_{co}$,$\lambda_{m}$,$\lambda_{e}$ and $\lambda_{v}$ are loss weights.

\subsection{Online Stage for Joint Model Fitting}
\label{sec:3-3}

\subsubsection{Compositive Volume Rendering}
Through the offline stage, we use the shape and appearance priors of hands and objects to build neural models, and then we fix the parameters of these models and optimize the poses of hands and objects in the hand-object interaction scene at online stage (Fig.~\ref{fig:fitting}).
Given sparse-view images, we first use multi-view-based pose estimation methods~\cite{learnable-triangulation,ghpt,cosypose} to obtain hand and object poses as initialization.
For each sampling point $\mathbf{x}$, it passes through hand and object models with initial poses to obtain $[\rho^{h}, \mathbf{c}^{h}]$ for hand and $[\rho^{o}, \mathbf{c}^{o}]$ for object, respectively. 
Then the rendering color and the foreground mask can be defined as:
\begin{align}
    &\hat{C}=\sum^{N}_{i=1} (T_{i}\alpha^{h}_{i} \mathbf{c}^{h}_{i} + T_{i}\alpha^{o}_{i} \mathbf{c}^{o}_{i}),\hat{M}=\sum^{N}_{i=1} (T_{i}\alpha^{h}_{i} + T_{i}\alpha^{o}_{i}),\nonumber\\
    &T_i=\exp\left(-\sum^{i-1}_{j}(\rho^{h}(j) + \rho^{o}(j))\Delta t_{j} \right),
\end{align}
where $\alpha_i=1-\exp(-\rho(i)\Delta t_{i})$, $\Delta t$ is the sampling distance between adjacent points along the ray, and $N$ is the number of sampling points along each ray.
\subsubsection{Single-Frame Based Loss Function}
We use render loss and pose regularizer loss to stabilize the positions of objects and hands, and adopt the interaction loss via SDF representation to avoid the penetration and encourage tight hand-object regions to be contact. 
Therefore, the loss function in the joint model fitting on single frame can be formulated as:
\begin{equation}
    \mathcal{L}_{fit}=\mathcal{L}_{render}+ \mathcal{L}_{pose}  + \mathcal{L}_{interact}.
  \label{fitting_loss}
\end{equation}

\vspace{1mm}
\noindent \textbf{Render Loss.}
The render loss consists of color and mask loss:  $\mathcal{L}_{render}=\lambda_{co} \mathcal{L}_{color}+\lambda_{m} \mathcal{L}_{mask}$.

\vspace{1mm}
\noindent \textbf{Pose Regularizer Loss.}
Inspired by~\cite{a-nerf}, we enforce the refined pose to be similar to the initial pose as: $\mathcal{L}_{pose}= \lambda_{h} \Vert \mathbf{J} - \mathbf{\hat{J}} \Vert_{2} + \lambda_{o} \Vert \mathbf{V} - \mathbf{\hat{V}} \Vert_{2}$,
where $\mathbf{J}$ and $\mathbf{V}$ are the refined 3D hand joints and object vertices, the $\mathbf{\hat{J}}$ and $\mathbf{\hat{V}}$ are the estimations of hand poses and object vertices as initialization, and $\lambda_{h}$ and $\lambda_{o}$ are loss weights. 

\vspace{1mm}
\noindent \textbf{Interaction Loss.}
The interaction loss includes penetration loss for solving the penetration and contact loss for forming reasonable contacts, which is defined as $\mathcal{L}_{interact}=\lambda_{p} \mathcal{L}_{p}+\lambda_{c} \mathcal{L}_{c}$, 
where $\mathcal{L}_{p}$ and $\mathcal{L}_{c}$ are the penetration loss and the contact loss, and $\lambda_{p}$ and $\lambda_{c}$ are loss weights.

For a sampling point $\mathbf{x}$, we calculate its SDF values for hand model $f_{hand}^{s}(\mathbf{x})$ and object model $f_{obj}^{s}(\mathbf{x})$, and penalize the SDF values of points, both the SDF values are negative, to become zero.
The penetration loss $\mathcal{L}_{p}$ can be formulated as: $\mathcal{L}_{p}=\frac{1}{\lvert N_{in} \rvert} \sum_{\mathbf{x}\in N_{in}}-(f_{hand}^{s}(\mathbf{x}) + f_{obj}^{s}(\mathbf{x}))$,
where $N_{in}$ represents the points whose SDF values for both hand model and object model are negative.

In order to encourage reasonable contact between hand and object, we use contact loss $\mathcal{L}_{c}$ to enforce the tight  hand-object regions to contact: $\mathcal{L}_{c}=\frac{1}{\lvert N_{c} \rvert}\sum_{\mathbf{x}\in N_{c}}( \lvert f_{hand}^{s}(\mathbf{x})\rvert + \lvert f_{obj}^{s}(\mathbf{x}) \rvert)$,
where $N_{c}$ represents the points with $\lvert f_{hand}^{s}(\mathbf{x})\rvert + \lvert f_{obj}^{s}(\mathbf{x}) \rvert < \varepsilon$ ($\varepsilon$ is set to 0.01).

\subsubsection{Video-Based Loss Function}
During online stage, our method can be also used for video sequences. 
We add smooth loss $\mathcal{L}_{smo}$ to reduce the pose jitters between frames. In order to make the contact area more stable and reduce sliding between hand and object, we propose a new stable contact loss $\mathcal{L}_{sta}$. Therefore, the loss function in the joint model fitting for video sequences can be formulated as:
\begin{equation}
    \mathcal{L}_{fit,video}=\mathcal{L}_{fit} + \mathcal{L}_{smo} + \mathcal{L}_{sta}.
  \label{fitting_loss_video}
\end{equation}

\vspace{1mm}
\noindent \textbf{Smooth Loss.}
The smooth loss encourages velocity of hand joints and object vertices to change smoothly:
$\mathcal{L}_{smo}=\frac{1}{N_t-1}\sum^{N_t-1}_{i=1}\mu_{h}\Vert \mathbf{J}_{i+1}-\mathbf{J}_{i} \Vert_{2} + \mu_{o}\Vert \mathbf{V}_{i+1}-\mathbf{V}_{i} \Vert_{2}$,
where $N_{t}$ is the frame number, and $\mu_{h}$ and $\mu_{o}$ are loss weights. 
\begin{figure}[t]
\centering%
\includegraphics[width=\linewidth]{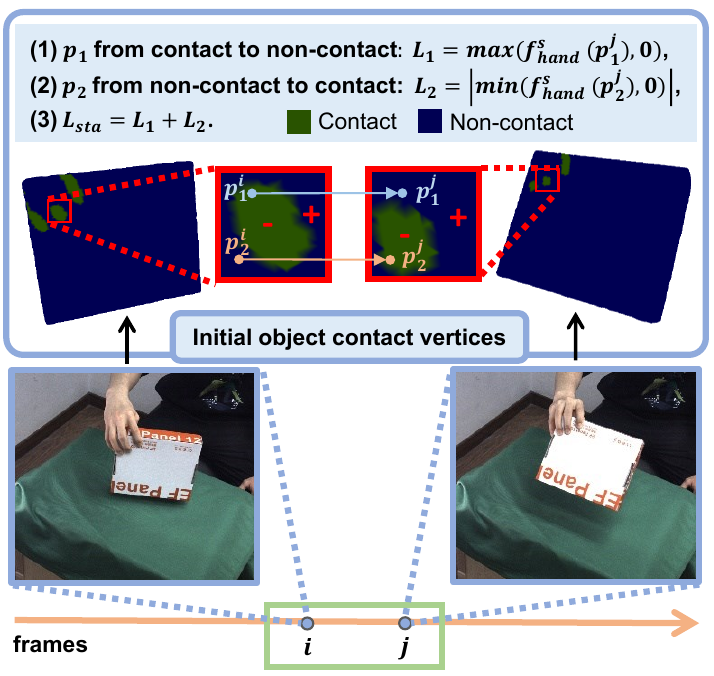}
\caption{Illustration of the stable contact loss between two frames. We extract the initial contact vertices on the object, and then penalize the inconsistent contact (i.e. contact to non-contact, non-contact to contact) of the object vertices between two frames.}
\label{fig:stablecontact}
\end{figure}

\vspace{1mm}
\noindent \textbf{Stable Contact Loss.}
Although the above loss functions are effective to get reasonable hand-object interaction results, 
there are still potential challenges. For example,
the contact loss $\mathcal{L}_c$ can effectively encourage contact for each frame, yet there is sliding between the hand and the object in the contact area.
To address this issue, we design a new stable contact loss to ensure reasonable and realistic hand-object contact for input sequences (Fig.~\ref{fig:stablecontact}).
We fix mesh models for each object pre-built in offline stage, and extract the initial contact vertices on the mesh, then we penalize the inconsistent contact of the vertices between frames. 

For frame $i$, we collect the initial contact vertices $\mathbf{p}_{1}^{i}=\mathbf{R}^{i}_{o}\mathbf{x}_{n}^{i}+\mathbf{T}^{i}_{o}$ on object surface, whose SDF value under the hand model $f_{hand}^{s}(\mathbf{p}_{1}^{i})$ is negative, where $\mathbf{x}_{n}^i$ represents the contact vertices in object model coordinate.
Then the initial contact vertices are transformed to the other frames using $\mathbf{p}_{1}^{j}=\mathbf{R}^{j}_{o}\mathbf{x}_{n}^{i}+\mathbf{T}^{j}_{o}$, and penalize the vertices whose $f_{hand}^{s}(\mathbf{p}_{1}^{j})$ are positive (i.e. contact to non-contact) using the loss $\mathcal{L}_{1}=max(f_{hand}^{s}(\mathbf{p}_{1}^{j}),0)$.
In order to avoid the degeneracy of contact (i.e., a wrongly predicted contact vertex of a frame could make its transformations in other frames to be contacted), we also need to refrain the non-contact object vertices from contacting with hand in other frames. 
Specifically, we query the nearest non-contact object vertex $d(\mathbf{x}_{n}^{i})$ to each contact point in frame $i$, where $d(\cdot)$ is a function to locate the closest object vertex to $\mathbf{x}_{n}^{i}$ that is not contact with hand. 
Then transform it to other frames $\mathbf{p}_{2}^{j} = \mathbf{R}_{o}^{j} d(\mathbf{x}_{n}^{i}) + \mathbf{T}_{o}^{j}$, and penalize the vertices with negative SDF values for hand model $f_{hand}^{s}(\mathbf{p}_{2}^{j})$ (i.e. non-contact to contact) using the loss $\mathcal{L}_{2}=\lvert min(f_{hand}^{s}(\mathbf{p}_2^{j}),0)\rvert$. Therefore, the stable contact loss can be formulated as:
\begin{equation}
     \mathcal{L}_{sta}=\frac{1}{M} \sum^{|N_{s}|}_{i=1}\sum^{|N_s|}_{j\neq i}(\mu_{si} \mathcal{L}_{1}+\mu_{so}\mathcal{L}_{2}),
\end{equation}
where ${N_s}$ represents the frames in which the number of contact points $x_{n}$ is greater than zero, $M$ is the total number of frame pairs with object contact among the sequence, and $\mu_{si}$ and $\mu_{so}$ are loss weights.

\section{Experiment}
\label{sec:exp}
\subsection{Datasets and Evaluation Metrics}
\label{sec:4-1}

\vspace{1mm}
\noindent \textbf{HandObject.}
Since there is no real-world dataset that can be used to train and evaluate our model, we use a multi-camera system with 8 cameras to collect a hand-object interaction dataset named HandObject.
It contains 85k images with the resolution of $266 \times 230$.

\vspace{1mm}
\noindent \textbf{Synthetic DexYCB.}
There are no images taken only for hand and object in the original DexYCB~\cite{chao2021dexycb}, which makes it impossible to train our offline model.
So we utilize DexYCB to generate a synthetic dataset named Synthetic DexYCB.
We use Pytorch3d~\cite{ravi2020pytorch3d} to render images of $400 \times 400$ and use the parametric hand texture model HTML~\cite{HTML} to add texture on hands. 

\begin{figure*}[ht]
\centering%
\includegraphics[width=\linewidth]{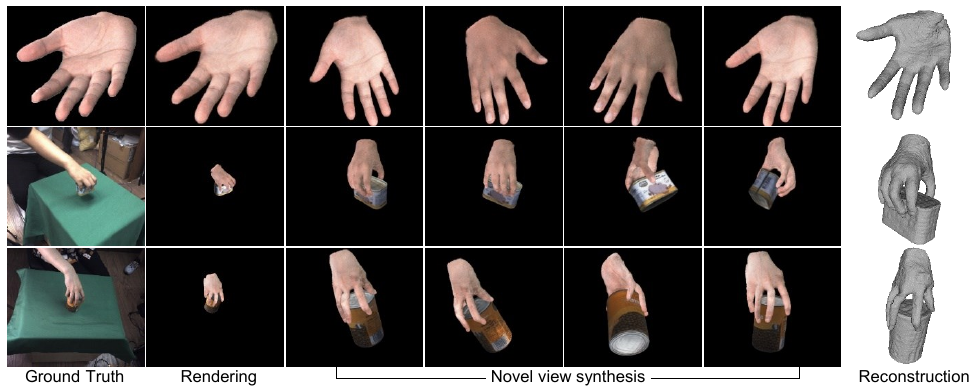}
\caption{Novel view synthesis and reconstruction of hand and hand-object interaction scenes.}
\label{fig:novel_view_syn_recon}
\end{figure*}

\vspace{1mm}
\noindent \textbf{Evaluation Metrics.}
1) We use the mean per joint position error (MPJPE) to measure the accuracy of hand skeleton pose. We follow~\cite{hinterstoisser2012admetric} to use the average distance (ADD), average closest point distance (ADD-S)~\cite{posecnn} and the value of average distance (AD) for evaluation of object pose estimation. We set ADD and ADD-S to be 15mm.
2) Rendering quality is evaluated with PSNR, SSIM and LPIPS metrics as~\cite{Nerf}. 
3) We follow~\cite{hasson2019learning} to use Penetration depth (mm) and Intersection volume (cm$^{3}$) to evaluate the interpenetration level.
4) To evaluate the effectiveness of smooth loss, we use Acceleration error ($mm/s^{2}$)~\cite{acc_err} to measure the average difference between ground truth 3D acceleration and predicted 3D acceleration of each hand joint (Acc-J) and each object vertex (Acc-V) respectively.  
To evaluate the effectiveness of our stable contact loss, we use the percentage of contact points IoU (PCI) as a new metric. We fix the mesh model for each object, and calculate PCI by averaging IoU of the contact vertices between adjacent frames. 

\vspace{1mm}
\noindent \textbf{Implementation Details.}
During offline stage, 
we train hand/object models with a single GeForce RTX 3080 Ti, costing 20 hours with 11.0 GB memory and 8 hours with 6.0 GB memory, respectively. 
We set $\lambda_{co}$, $\lambda_{m}$, $\lambda_{e}$ to 1, and 
set $\lambda_{v}$ to increment from 0 to 1 after 10k iterations and then keep it at 1 in Eq.~\ref{eq17}.
We form a training batch by randomly sampling 441 rays from an image, with 64 coarse sampling points and 64 fine sampling points.
In the single-frame model fitting part of the online stage~(Sec.~\ref{sec:3-3}), we first use render loss and pose regularizer loss to iterate over each image 30 times, and the loss weights $\lambda_{co}$, $\lambda_{m}$, $\lambda_{h}$, and $\lambda_{o}$ are set to 1, 0.5, 100, and 5. Then the interaction loss is added and iterates 25 times, where $\lambda_{h}$, $\lambda_{o}$, $\lambda_{p}$, and $\lambda_{c}$ are set to 30, 20, 20 and 30. 
We use the single-frame optimized pose as the initialization, and then add smooth loss and stable contact loss for video-based model fitting. The loss weight $\lambda_{co}, \lambda_{m}, \mu_{h}, \mu_{o}, \mu_{si}, \mu_{so}$ are set to 0.5, 0.25, 50, 50, 100 and 5. During the online stage, we sample 64 coarse sampling points and 128 fine sampling points on a ray. The optimization and rendering times for one frame are 3 minutes and 30 seconds, respectively.  

\begin{table}[h]
\centering
\resizebox{\linewidth}{!}{
\begin{tabular}{c|c|c|c|c}
\hline
{Method} &MPJPE $\downarrow$ &AD $\downarrow$&ADD $\uparrow$&ADD-S $\uparrow$\\
\hline
CosyPose (CP) ~\cite{cosypose}&-&21.10&60.46&84.20\\
I2L~\cite{i2l}&18.39&-&-&-\\
I2L+CP+Mesh Fitting&20.60&21.10&60.54&84.25\\
GHPT~\cite{ghpt}&13.28&-&-&-\\
GHPT+CP+Ours&10.80&\textbf{15.78}&\textbf{71.09}&\textbf{93.67}\\
LT~\cite{learnable-triangulation}&9.66&-&-&-\\
LT+CP+A-NeRF&9.22&20.94&61.01&84.39\\
LT+CP+Ours&\textbf{9.09}&15.95&70.73&93.20\\
\hline
\end{tabular}
}
\caption{Comparison on pose estimaion with SoTA methods. High-quality rendering results with our method are conducive to obtaining more accurate poses in the rendering-based optimization. 
}
\label{table:pose_compare}
\end{table}

\begin{figure*}[ht]
\centering%
\includegraphics[width=\linewidth]{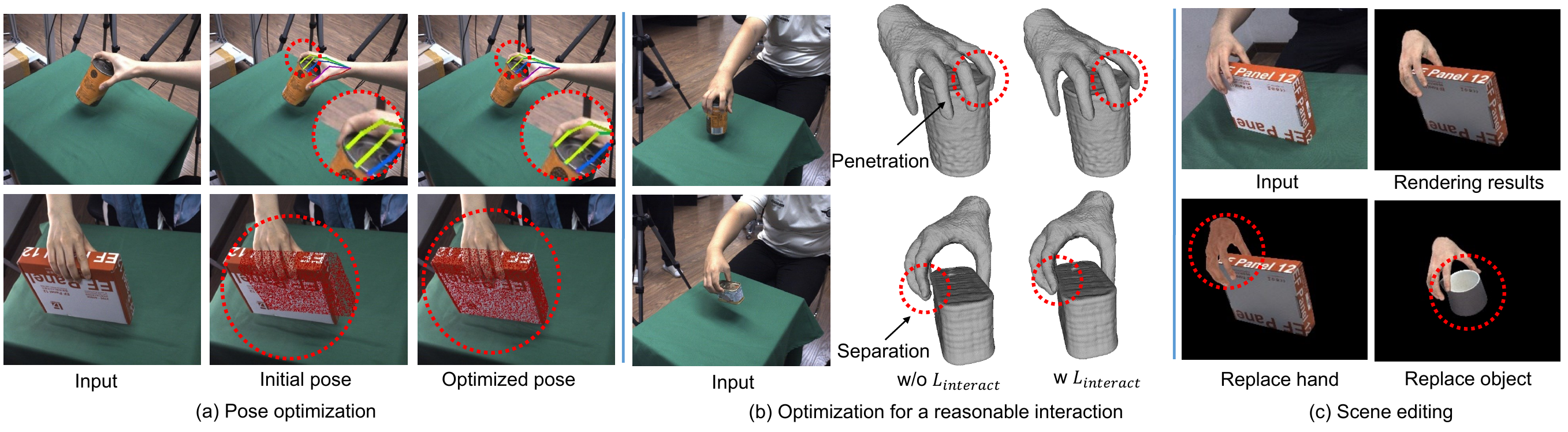}
\caption{(a) Effect of pose optimization in online stage with joint model fitting. Optimization can improve the pose accuracy. (b) Effect of interaction loss. Interaction loss can  facilitate to achieve reasonable hand-object interaction. (c) Editing of hand-object interaction scenes. We can replace the hand, object models and change the poses to get realistic rendering results.}
\label{fig:pose_interaction_edit}
\end{figure*}
\begin{figure*}[ht]
\centering%
\includegraphics[width=\linewidth]{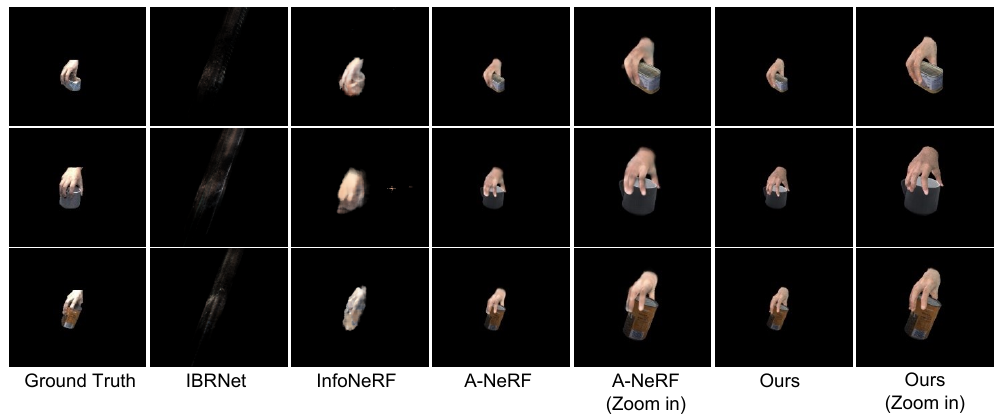}
\caption{Rendering quality comparison on the HandObject dataset. We zoom in the rendering results for demonstration. Our method can achieve high-quality rendering results. Compared with A-NeRF~\cite{a-nerf} based hand model, our method preserves more realistic details.}
\label{fig:render_quality}
\end{figure*}

\begin{table*}[ht]
\centering
\resizebox{\linewidth}{!}{
\begin{tabular}{cc|c|c|c|c|c|c|c|c|c|c|c|c|c|c|c|c|c|c}
\hline
\multirow{3}{*}{ }&\multirow{3}{*}{Object}&  \multicolumn {6}{c|}{8 views} & \multicolumn {6}{c|}{6 views} & \multicolumn {6}{c}{3 views} \\
\cline{3-20}
&&\multicolumn {3}{c|}{CosyPose~\cite{cosypose}} & \multicolumn {3}{c|}{Ours}& \multicolumn {3}{c|}{CosyPose~\cite{cosypose}} & \multicolumn {3}{c|}{Ours}& \multicolumn {3}{c|}{CosyPose~\cite{cosypose}} & \multicolumn {3}{c}{Ours}\\
\cline{3-20}
&&AD $\downarrow$& ADD $\uparrow$ & ADD-S $\uparrow$ & AD $\downarrow$ & ADD $\uparrow$ & ADD-S $\uparrow$&AD $\downarrow$& ADD $\uparrow$ & ADD-S $\uparrow$ & AD $\downarrow$ & ADD $\uparrow$ & ADD-S $\uparrow$&AD $\downarrow$& ADD $\uparrow$ & ADD-S $\uparrow$ & AD $\downarrow$ & ADD $\uparrow$ & ADD-S $\uparrow$\\
\hline
\multirow{5}{*}{\rotatebox{90}{\begin{tabular}[c]{@{}c@{}}Synthetic\\ DexYCB\end{tabular}}}&

002-master-chef-can & 10.40 &77.48 & 100.00 & \textbf{5.56} &\textbf{90.73} &100.00  & 12.48  & 65.56 &  100.00  &\textbf{8.23} &\textbf{80.79} &100.00  & 10.71  & 90.07  & 100.00 &\textbf{7.08} &\textbf{95.36} &100.00\\
&
003-cracker-box & 15.15 &  70.95 &  89.86 & \textbf{3.57} &\textbf{97.97} &\textbf{100.00}  & 16.98  & 60.81 &  91.89 &\textbf{5.94} &\textbf{90.54} &\textbf{100.00}  & 15.98  & 72.97  & 89.86  &\textbf{6.73}  &\textbf{93.24} &\textbf{97.30}\\
&
006-mustard-bottle & 45.10  & 6.54  &  73.20 &\textbf{31.93} &\textbf{53.59} &\textbf{97.39} & 48.81 &  13.73  & 60.13 &\textbf{35.63} &\textbf{43.14} &\textbf{87.58}  & 45.99 &  7.84  & 67.97&\textbf{34.37} &\textbf{38.56} &\textbf{91.50}\\
&
010-potted-meat-can & 27.58  & 23.02 &  90.65 &\textbf{11.38} &\textbf{76.98} &\textbf{100.00}  & 37.47  & 8.63  &  84.89 &\textbf{26.95} & \textbf{48.92}& \textbf{99.28} & 28.27 & 28.06  & 78.42 &\textbf{15.88} &\textbf{69.78} &\textbf{99.28}\\
&
011-banana & 29.29 & 36.55 &  60.00 &\textbf{13.53} &\textbf{69.66} &\textbf{95.86} & 33.91  & 26.90 & 51.72 & \textbf{19.17} & \textbf{45.52} & \textbf{87.59}  & 46.85 &  22.07  & 42.76  &\textbf{44.01} &\textbf{37.24} &\textbf{73.10}\\
\hline\hline
\multirow{4}{*}{\rotatebox{90}{\begin{tabular}[c]{@{}c@{}}Hand-\\ Object\end{tabular}}}&
bean-can & 19.43  & 62.65 &  90.62  &\textbf{16.86}&\textbf{66.83} & \textbf{93.80}& 23.84  & 49.92  & 87.10 & \textbf{20.91}  &\textbf{57.45}  &\textbf{91.12}  & 21.46  & 58.63  & 87.94  &\textbf{21.06} &\textbf{59.30} & \textbf{88.44}\\
&
box & 20.59  & 59.39 &  87.43 &\textbf{14.95} &\textbf{76.55} &\textbf{95.57}  & 23.47 &  51.65  & 80.10 & \textbf{18.35}  &\textbf{66.16}&\textbf{91.54} & 35.44  & 26.35  & 53.26 &\textbf{32.75} &\textbf{33.28} &\textbf{59.31}\\
&
cup & 25.72 &  52.07  & 74.01  &\textbf{17.79}  &\textbf{61.76} &\textbf{90.13}  & 18.10  & 60.53  & 92.25  & \textbf{14.18} &\textbf{76.04}  &\textbf{96.74}  & 27.19 & 33.92& 73.39 &\textbf{24.76} &\textbf{39.03}&\textbf{79.82}\\ 
&
meat-can & 15.35 & 75.64 & 90.13 & \textbf{13.74} & \textbf{79.14} & \textbf{93.47} & 15.21 &  71.97  & 92.52   & \textbf{13.71} & \textbf{77.23}  & \textbf{94.11} &  21.03  & 47.29  & 86.46 &\textbf{19.56} &\textbf{54.14} &\textbf{89.49}\\
\hline
\end{tabular}
}
\caption{Comparison of object pose estimation under different camera views on HandObject and Synthetic DexYCB.
}
\label{table:obj_pose_estimation}
\end{table*}

\subsection{Novel View Synthesis and Reconstruction}
We show novel view synthesis and reconstruction of hand and hand-object interaction  in Fig.~\ref{fig:novel_view_syn_recon}. Our models contain realistic appearance and geometry details and can achieve full 360 degree free-viewpoint rendering.

\subsection{Comparison to State-of-the-art Methods}
\label{sec:4-2}
\vspace{1mm}
\noindent\textbf{Pose Estimation.}
In hand pose estimation, we compare with the state-of-the-art (SoTA) multi-view pose estimation methods including LT~\cite{learnable-triangulation} and GHPT~\cite{ghpt}, a single view pose estimation method I2L~\cite{i2l} and a fitting method based on A-NeRF~\cite{a-nerf}.
In object 6D pose estimation, we compare with the SoTA multi-view method CosyPose~\cite{cosypose}.
During comparison, we optimize the pose initialized by LT, GHPT and CosyPose, and replace our hand model with A-NeRF based hand model for pose refinement (i.e. 'LT+CP+A-NeRF').
We also compare with the mesh-based fitting methods. I2L predicts the MANO parameters, and we use the untextured MANO hand mesh and the object mesh obtained in the offline stage to optimize the pose by fitting without color loss (i.e. 'I2L+CP+Mesh Fitting').
We show the results on HandObject under eight views in Table~\ref{table:pose_compare}.
We also test the accuracy of pose estimation under different number views on HandObject and Synthetic DexYCB.
Table~\ref{table:hand_pose_multiview} shows the hand pose estimation results compared with LT, and the qualitative comparisons are demonstrated in the first row of Fig.~\ref{fig:pose_interaction_edit}(a).
Table~\ref{table:obj_pose_estimation} shows object pose estimation results compared with CosyPose, and the qualitative comparisons are shown in the second row of Fig.~\ref{fig:pose_interaction_edit}(a). 
We observe that our method outperforms the SoTA methods. Benefiting from the pre-built models, our method can exploit more dense supervision on image pixels than sparse keypoint supervision in LT, I2L, GHPT and CosyPose. Compared with A-NeRF hand model based fitting, our high-quality rendering models are conducive to obtaining more accurate poses. Compared with untextured mesh based fitting, we find that it is inferior to the color loss to provide sufficient constraints to achieve accurate pose.

\begin{table}[h]
\centering
\resizebox{\linewidth}{!}{
\begin{tabular}{cc|c|c|c|c|c|c}
\hline
\multirow{2}{*}{ }&\multirow{2}{*}{Object} & \multicolumn {2}{c|}{8 views} & \multicolumn {2}{c|}{6 views}& \multicolumn {2}{c}{3 views}\\ \cline{3-8} 
&&LT~\cite{learnable-triangulation}&Ours&LT~\cite{learnable-triangulation}&Ours&LT~\cite{learnable-triangulation}&Ours\\
\hline
\multirow{5}{*}{\rotatebox{90}{\begin{tabular}[c]{@{}c@{}}Synthetic\\ DexYCB\end{tabular}}}&

002-master-chef-can & 9.61& \textbf{7.84} &10.08& \textbf{8.21} &15.57&\textbf{13.65}\\
&
003-cracker-box & 10.72 & \textbf{9.87} & 12.09 & \textbf{11.25} & 12.06 & \textbf{11.03}\\
&
006-mustard-bottle & 9.15 & \textbf{7.73} & 10.22 & \textbf{8.58} & 10.74 &\textbf{9.31}\\
&
010-potted-meat-can & 8.27 & \textbf{7.08} & 8.68 & \textbf{7.62} & 12.06 &\textbf{10.62}\\
&
011-banana & 11.21 & \textbf{10.53} & 11.67 &\textbf{10.98} & 13.50 &\textbf{13.47}\\
\hline\hline
\multirow{4}{*}{\rotatebox{90}{\begin{tabular}[c]{@{}c@{}}Hand-\\ Object\end{tabular}}}&
bean-can & 8.85 &\textbf{8.07} & 9.31 & \textbf{8.17} & 14.05 &\textbf{11.98}\\
&
box & 9.63& \textbf{9.29} & 10.25 & \textbf{9.89} & 17.81 &\textbf{16.73}\\
&
cup & 10.51 & \textbf{9.89} & 11.12 & \textbf{10.45} & 16.19 & \textbf{15.29}\\ 
&
meat-can & 8.97 & \textbf{8.20} & 9.46 & \textbf{8.50} & 15.33 &\textbf{13.50}\\
\hline
\end{tabular}
}
\caption{MPJPE (mm) of hand pose estimation under different view numbers on HandObject and Synthetic DexYCB.
}
\label{table:hand_pose_multiview}
\end{table}

\begin{table}[ht]
\centering
\resizebox{\linewidth}{!}{
\begin{tabular}{c|c|c|c}
\hline
{Method} &PSNR $\uparrow$& SSIM $\uparrow$ &LPIPS $\downarrow$  \\
\hline
IBRNet~\cite{ibrnet} & 17.02 & 73.75 & 0.291 \\
InfoNeRF~\cite{infonerf} & 18.70 & 87.92 & 0.161 \\
A-NeRF~\cite{a-nerf} & 21.99 & 93.40 & 0.066 \\
Ours & \textbf{22.20}&\textbf{93.71}&\textbf{0.059} \\
\hline
\end{tabular}
}
\caption{Quantitative comparison of rendering quality. }
\label{table:result_synthesis}
\end{table}

\begin{table*}[ht]
\centering
\centering
\resizebox{\linewidth}{!}{
\begin{tabular}{cc|c|c|c|c|c|c|c|c}
\hline
\multirow{2}{*}{ }&\multirow{2}{*}{Object} & \multicolumn {4}{c|}{$\mathcal{L}_{render}+ \mathcal{L}_{pose}$} & \multicolumn {4}{c}{$\mathcal{L}_{render}+\mathcal{L}_{pose} +\mathcal{L}_{interact}$}\\\cline{3-10}
&&MPJPE $\downarrow$& AD $\downarrow$ & Int-Vol $\downarrow$ & Pen-Dep $\downarrow$ &MPJPE$\downarrow$&  AD $\downarrow$ &Int-Vol $\downarrow$ & Pen-Dep $\downarrow$\\
\hline
\multirow{5}{*}{\rotatebox{90}{\begin{tabular}[c]{@{}c@{}}Synthetic\\ DexYCB\end{tabular}}}&

002-master-chef-can & 8.60 & 6.06 &8.54  &9.65  &\textbf{7.84}&\textbf{5.56} &\textbf{4.87} &\textbf{6.09}  \\
&
003-cracker-box & 9.97 & 4.43 &5.84 &13.09 &\textbf{9.87}&\textbf{3.57}&\textbf{3.60} &\textbf{6.36} \\
&
006-mustard-bottle & 8.16&32.71 &6.68 &8.73 &\textbf{7.73}&\textbf{31.93}&\textbf{2.07} &\textbf{3.08} \\
&
010-potted-meat-can & 7.19&12.09 &4.39 &9.22 &\textbf{7.08}&\textbf{11.38}&\textbf{1.58} &\textbf{1.88} \\
&
011-banana & 10.71&14.19 &2.17& 5.34&\textbf{10.53} &\textbf{13.53}&\textbf{0.93} &\textbf{1.54} \\
\hline\hline
\multirow{4}{*}{\rotatebox{90}{\begin{tabular}[c]{@{}c@{}}Hand-\\ Object\end{tabular}}}&
bean-can & \textbf{7.72} & 16.92 &4.95 & 6.28 &8.07& \textbf{16.86} &\textbf{3.17} &\textbf{3.52} \\
&
box & \textbf{9.23}&15.18& 6.91  & 11.12 &9.29 &\textbf{14.95}&\textbf{5.79} & \textbf{8.68} \\
&
cup & \textbf{9.77} & 17.89 & 5.00 & 9.14 &9.89 &\textbf{17.79}&\textbf{4.26} &\textbf{6.82} \\ 
&
meat-can & \textbf{7.90} &13.81 & 4.26& 6.50&8.20 &\textbf{13.74}&\textbf{3.23} & \textbf{3.85} \\
\hline
\end{tabular}
}
\caption{Effect of interaction loss on Interpenetration level.}
\label{table:pen_con_res}
\end{table*}

\begin{table*}[ht]
\begin{center}
\resizebox{\linewidth}{!}{
\begin{tabular}{c|c|c|c|c|c|c|c|c|c}
\hline
\multirow{2}{*}{Datasets} & \multicolumn {3}{c|}{$\mathcal{L}_{fit}$} & \multicolumn {3}{c|}{$\mathcal{L}_{fit}+\mathcal{L}_{smo}$}& \multicolumn {3}{c}{$\mathcal{L}_{fit}+\mathcal{L}_{smo}+\mathcal{L}_{stable}$}\\ \cline{2-10} 
&Acc-J$\downarrow$&Acc-V$\downarrow$&PCI$\uparrow$&Acc-J$\downarrow$&Acc-V$\downarrow$&PCI$\uparrow$&Acc-J$\downarrow$&Acc-V$\downarrow$&PCI$\uparrow$\\
\hline
Synthetic DexYCB &8.04&26.76&10.98&\textbf{6.78}&19.51&17.38&7.51&\textbf{19.18}&\textbf{35.02}\\

\hline\hline
HandObject&6.28&10.07&29.68&\textbf{6.20}&6.35&36.37&6.25&\textbf{6.10}&\textbf{51.18}\\

\hline
\end{tabular}
}
\end{center}
\caption{Effect of smooth loss and stable contact loss. Smooth loss will reduce pose jitters with lower acceleration error, and stable contact loss will make the contact area more stable with higher PCI.}
\label{table:smooth_stable}
\end{table*}

\vspace{1mm}
\noindent\textbf{Rendering Quality.}
We compare the rendering quality in hand-object interaction scenes with A-NeRF~\cite{a-nerf}, IBRNet~\cite{ibrnet} and InfoNeRF~\cite{infonerf} on HandObject dataset under five test views. 
Table~\ref{table:result_synthesis} shows quantitative comparison, and Fig.~\ref{fig:render_quality} shows qualitative comparison. 
We replace our hand model with A-NeRF based hand model and use the same object model for fitting and rendering (i.e. 'A-NeRF').
The rendering results with our method perform better than others,  
because our offline models provide strong shape and appearance priors which are more suitable for few-shot neural rendering. Compared to the density representation in A-NeRF, the SDF representation in our model makes the shape sharper.

\subsection{Ablation Study}
\label{sec:4-4}
\vspace{1mm}
\noindent \textbf{Effect of Interaction Loss.}
We compare the interpenetration level as shown in Table~\ref{table:pen_con_res} and Fig.~\ref{fig:pose_interaction_edit}(b).
We observe that our model with interaction loss can achieve excellent performance on Penetration depth (Pen-Dep) and Intersection volume (Int-Vol), i.e., while only sacrificing negligible performance drop in hand pose. 
We also compare the rendering quality on the HandObject dataset as shown in Table~\ref{table:render_qulity_dif_loss}. After fitting (i.e. '$\mathcal{L}_{render}+ \mathcal{L}_{pose}$', '$\mathcal{L}_{render}+\mathcal{L}_{pose} +\mathcal{L}_{interact}$'), the rendering quality becomes better. The interaction loss $\mathcal{L}_{interact}$ can further improve 
the rendering quality, because the incorrect  color caused by unreasonable interactions such as penetration can be reduced by the loss.

\begin{table}[ht]
\centering
\resizebox{\linewidth}{!}{
\begin{tabular}{c|c|c|c}
\hline
Method & PSNR  $\uparrow$& SSIM  $\uparrow$& LPIPS $\downarrow$\\\cline{1-4}
w/o $\mathcal{L}_{fit}$  & 22.07 & 93.32 & 0.0622 \\
$\mathcal{L}_{render}+ \mathcal{L}_{pose}$ & 22.17 & 93.54 & 0.0607 \\ 
$\mathcal{L}_{render}+\mathcal{L}_{pose} +\mathcal{L}_{interact}$ & \textbf{22.25} & \textbf{93.57} & \textbf{0.0605}\\

\hline
\end{tabular}
}
\caption{Effect of interaction loss on rendering quality. }
\label{table:render_qulity_dif_loss}
\end{table}

\vspace{1mm}
\noindent \textbf{Effect of Smooth Loss and Stable Contact Loss.}
Table~\ref{table:smooth_stable} shows the results on acceleration error and PCI. Compared with applying $\mathcal{L}_{fit}$ only, we observe that the smooth loss can reduce pose jitters and lead to smoother pose change with lower acceleration error.
After adding stable contact loss, the PCI values increase, indicating the contact regions tend to be stable, and acceleration error on object pose is also reduced, indicating a further reduction in object jitters.  

\subsection{Hand-Object Interaction Editing}
\label{sec:4-3}
Our model can be driven by controllable variables such as hand pose $\mathbf{J}$, object pose $\mathbf{R}_{o}$ and $\mathbf{T}_{o}$.
We can edit hand-object interaction scenes, including replacing hand or object models, and poses as shown in  Fig.~\ref{fig:pose_interaction_edit}(c).

\section{Conclusion}
\label{sec:con}
We propose a novel neural rendering and pose estimation system for hand-object interaction from sparse view images. 
We design a two-stage approach (i.e. offline model building and online model fitting) to achieve accurate hand-object pose estimation and photo-realistic novel view synthesis. We utilize effective geometric constraints to conduct rendering-based online model fitting.
Various experiments demonstrate that our method outperforms the SoTA pose estimation and  few-shot neural rendering methods.
In the future work, we will take into account lighting conditions to reduce unrealistic results caused by shadows from different illuminations and improve the efficiency of our method with Instant-NGP~\cite{instant_ngp}. 

\noindent \textbf{Acknowledgments.} This work was supported in part by National Key R\&D Program of China (2022ZD0117900).

{\small
\bibliographystyle{ieee_fullname}
\bibliography{egbib}
}

\clearpage

\setcounter{section}{0}
\renewcommand\thesection{\Alph{section}}
\setcounter{table}{0}
\renewcommand\thetable{\Alph{table}}
\setcounter{figure}{0}
\renewcommand\thefigure{\Alph{figure}}

\title{Novel-view Synthesis and Pose Estimation for Hand-Object Interaction\\ from Sparse Views \\
-Supplementary Material-}

\author{
Wentian Qu$^{1,2}$\qquad 
Zhaopeng Cui$^{3}$\qquad
Yinda Zhang$^{4}$\qquad
Chenyu Meng$^{1,2}$\qquad
Cuixia Ma$^{1,2}$\\
Xiaoming Deng$^{1,2}$\qquad
Hongan Wang$^{1,2}$$^*$\\
$^1$Institute of Software, Chinese Academy of Sciences \quad $^2$University of Chinese Academy of Sciences\\ 
$^3$State Key Lab of CAD$\&$CG, Zhejiang University  \quad $^4$Google
}
\maketitle

In this supplementary material, we first introduce the implementation details of our method and our dataset (Sec.~\ref{sec:supp_1}). Then we show more experimental results (Sec.~\ref{sec:supp_2}). Finally, we summarize several limitations of our method (Sec.~\ref{sec:supp_3}). More results can be found in the supplementary video.

\section{Method Details}
\label{sec:supp_1}

\subsection{Details of our HandObject Dataset}
\begin{figure*}[ht]
\centering%
\includegraphics[width=\linewidth]{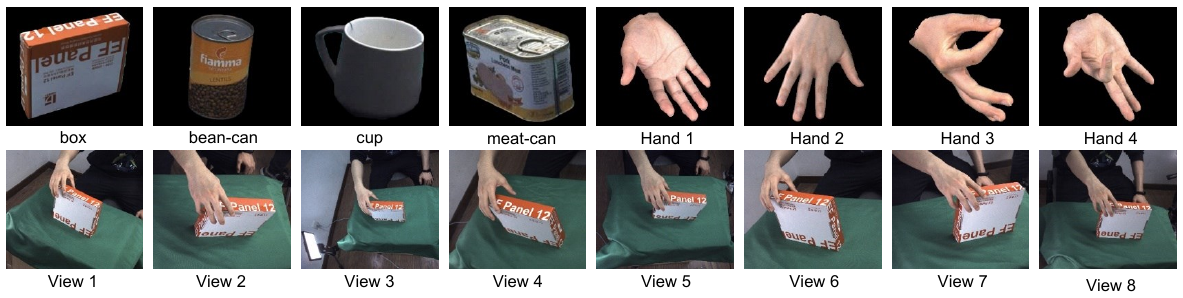}
\caption{HandObject dataset. The first row shows four objects and four hands. The second row shows eight perspectives in the hand-object interaction scenes.}
\label{fig:dataset}
\end{figure*}

Our HandObject dataset is collected with a sparse-view camera system with 8 calibrated color cameras.
We capture a collection of images, including hands of four persons, four objects (including 'bean-can', 'cup', 'box' and 'meat-can'), and hand-object interactions. Fig.~\ref{fig:dataset} shows hand and object instances, and the camera viewpoints of camera system.
We detect hand and object using hand and object detection method~\cite{hand_object_detection}, and use PointRend~\cite{pointrend} to obtain rough foreground masks.
We use MediaPipe~\cite{mediapipe} to obtain the initial hand skeleton pose of the hand images and use CosyPose~\cite{cosypose} to obtain the initial object pose in the offline stage. 

\begin{figure}[h]
\centering%
\includegraphics[width=\linewidth]{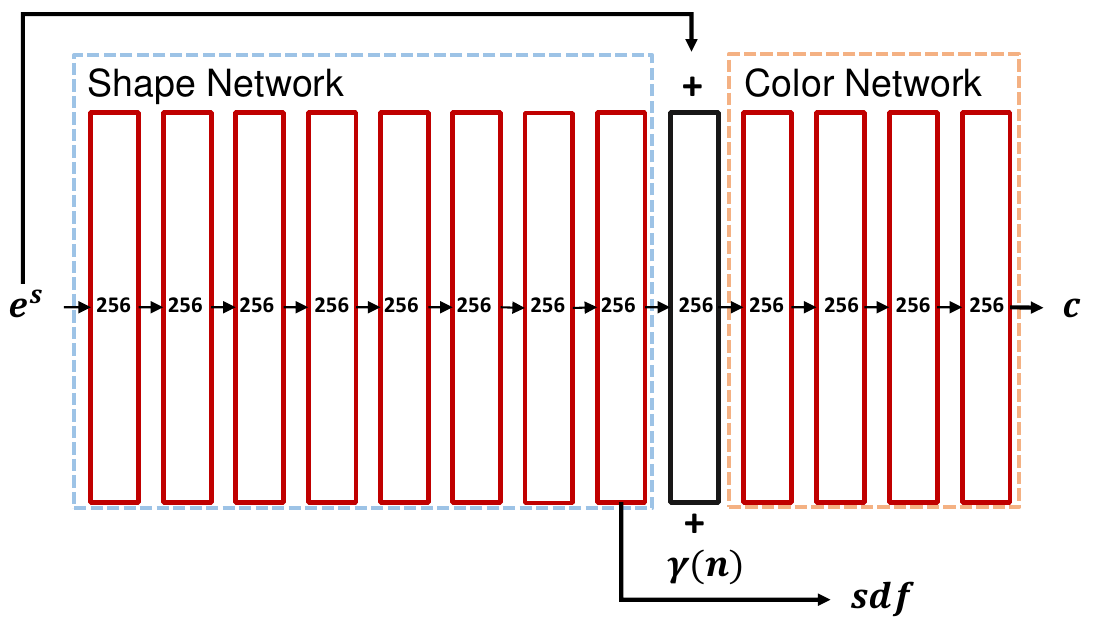}
\caption{The network architecture of our hand model.}
\label{fig:network}
\end{figure}

\subsection{Network Architecture}
In Sec.~3.1 of our main paper, we introduce our offline models, and we will show more details in this section.

\vspace{1mm} 
\noindent \textbf{Differences to A-NeRF~\cite{a-nerf}.}
Our hand network architecture is shown in Fig.~\ref{fig:network}, and our method has two key improvements compared to A-NeRF~\cite{a-nerf}. First, the embedding vectors of our hand model is different from A-NeRF. 
We add positional encoding $\lambda(\cdot)$ and cutoff on relative direction $\mathbf{r}$ of each sampling point $\mathbf{x}$ on camera ray. We define the cutoff point $\bar{\mathbf{v}}=$[0.08, 0.03, 0.03, 0.02, 0.02, 0.03, 0.02, 0.02, 0.02, 0.03, 0.02, 0.02, 0.02, 0.03, 0.02, 0.02, 0.02, 0.03, 0.02, 0.02, 0.02] by clustering analysis of bone length $\mathbf{L}$.
This modification allows the model to preserve better piece-wise rigidity of the fingers.
Second, we add positional encoding $\lambda(\cdot)$ on the normal $\mathbf{n}_{hand}$ of each sampling point $\mathbf{x}$, i.e. the derivation of the SDF value of sampling point, to replace the original appearance code used in A-NeRF. We encode the normal $\mathbf{n}_{hand}$ to get reliable color prediction in our hand model because the normal of a surface point is a key clue for geometric information \cite{Deng_2021_CVPR}. See Fig.~\ref{fig:supp_com_anerf} for qualitative comparison with A-NeRF. 

\vspace{1mm}  
\noindent \textbf{Network Details.}
Our network consists of shape network and color network. The shape network is an 8-layer MLP (width=256), taking as input embedding vectors $\mathbf{e}^{s}$ and output $sdf$. 
The color network is a 4-layer MLP (width=256). The color network of our hand model takes $\mathbf{e}^{s}$ combined with $\gamma(\mathbf{n})$ and the feature produced by shape network as input and output color $\mathbf{c}$. 
The color network of our object model takes $\mathbf{e}^{s}$ combined with $\gamma(\mathbf{n})$, the ray direction under object coordinate system $\mathbf{l}_{o}$ and the feature produced by shape network as input and output color $\mathbf{c}$. Our offline stage model needs 300k iterations, and we add VGG loss at the 90,000th iteration in Eq.5 of our main paper.

\subsection{Bone Transformation}
\label{sec:bonetrans}
In Sec.~3.1 of our main paper, we use bone transformation for coordinate transformation to canonical pose, and we further introduce the details of the bone transformation in this section.

The hand skeleton pose $\mathbf{J}=\{\mathbf{J}_{i}\}^{21}_{i=1} \in \mathbb{R}^{21\times3}$ can be decoupled as pose parameters $\theta \in \mathbb{R}^{36}$ and bone length $\mathbf{L}\in\mathbb{R}^{20}$.
We can also use pose parameters $\theta$ and bone length $\mathbf{L}$ to get a pose $\mathbf{J}$ by forward kinematics, and then use $\mathbf{J}$ to calculate the bone transformation $\mathbf{B}^{-1} \in \mathbb{R}^{21\times4\times4}$, which can convert the current pose $\mathbf{J}$ to the canonical pose $\mathbf{T}\in\mathbb{R}^{21\times3}$. Our bone transformation consists of a global transformation matrix $\mathbf{B}_{g}\in\mathbb{R}^{1\times4\times4}$ of the root joint to obtain 3D root-aligned joints, and a local transformation matrix $\mathbf{B}_{l}\in\mathbb{R}^{21\times4\times4}$ of  joints to convert the root-aligned joints to the canonical pose. The bone transformation can be defined as $\mathbf{B}^{-1}=\mathbf{B}_{l}\mathbf{B}_{g}$. In the following, we first introduce the rotation angles contained in the hand pose, and then introduce the key idea of the bone transformation calculation in Sec.~\ref{sec:globallocal}.

\begin{figure}[ht]
\centering%
\includegraphics[width=\linewidth]{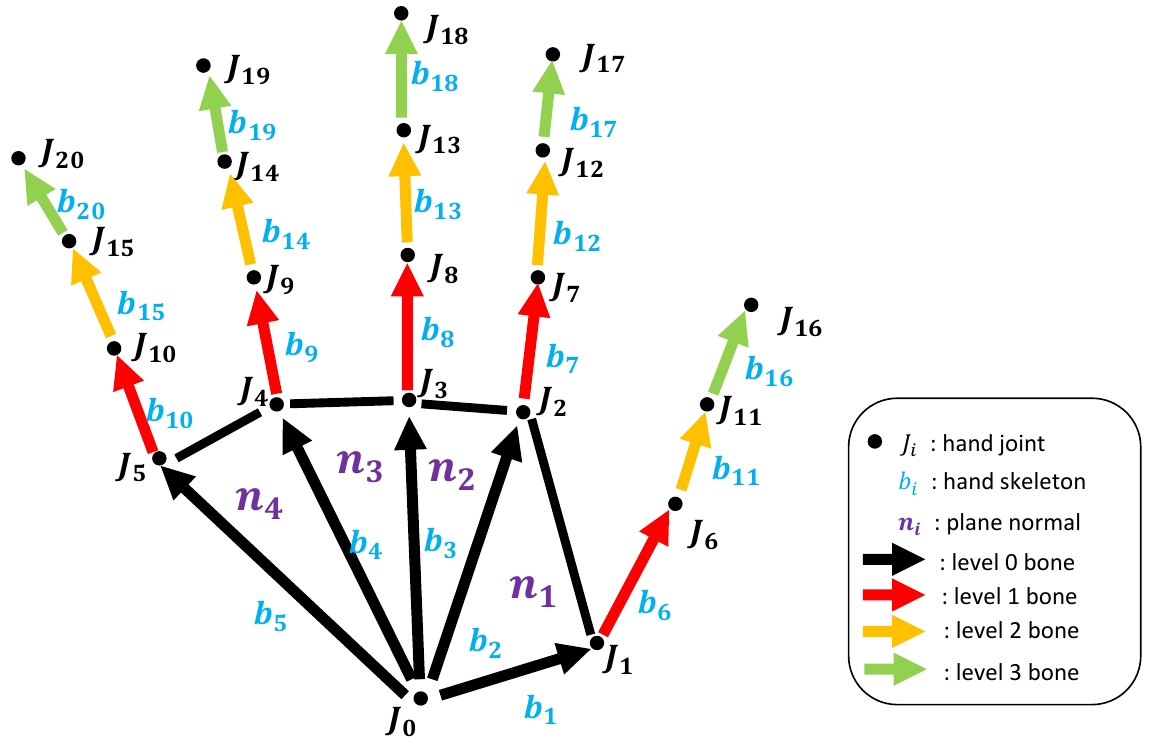}
\caption{Hand structure to get  bone transformation. In order to calculate bone transformation, we define four levels of bones, illustrated with different colors, black arrow as level-0 bone, red arrow as level-1 bone, yellow arrow as level-2 bone, green arrow as level-3 bone.}
\label{fig:sup_bt}
\end{figure}

We follow HALO~\cite{skeleton_hand} to define the structure of the right hand used for bone transformation calculation (see Fig.~\ref{fig:sup_bt}). 
The hand structure contains 21 joints $\mathbf{J}_{i}\in\mathbb{R}^{3}$, and 20 bones $\mathbf{b}_{i} \in \mathbb{R}^{3}$. The palm is divided into four planes (each plane passes three neighboring joints on the palm), and we define the normal direction of each plane as $\mathbf{n}_{i} \in \mathbb{R}^{3}$.  The hand joints are divided into four levels (see Fig.~\ref{fig:sup_bt}), and we can get 37 controllable angles, including the angles between the normal directions of two adjacent palm planes (3 in total), the angles between adjacent level-0 bones (4 in total), and level 1-3 bone has flexion angle and abduction angles in their respective local coordinate systems (30 in total). 

\subsection{Global and Local Transformation}
\label{sec:globallocal}
We follow HALO~\cite{skeleton_hand} to calculate the global transformation $\mathbf{B}_{g}$ and the local transformation $\mathbf{B}_{l}$ in Sec.~\ref{sec:bonetrans}. 

\vspace{1mm}  \noindent \textbf{Global Transformation.} The global transformation $\mathbf{B}_{g}$ includes the rotation $\mathbf{R}_{palm}\in\mathbb{R}^{3\times3}$ and translation $\mathbf{T}_{palm}\in\mathbb{R}^{3}$ of the root joint $\mathbf{J}_{0}$. First, we translate $\mathbf{J}_{0}$ to the origin $\mathbf{O}$ of the world coordinate, align $\mathbf{b}_{3}$ with $-\mathbf{Y}$ axis, and then align the plane passing $\mathbf{b}_{2}$ and $\mathbf{b}_{3}$ with $\mathbf{X}-\mathbf{O}-\mathbf{Y}$ plane to achieve global alignment. Finally, we get the global matrix $\mathbf{B}_{g}$ by combining $\mathbf{R}_{palm}$ and $\mathbf{T}_{palm}$. 

\vspace{1mm}  \noindent \textbf{Local Transformation.}
We follow HALO~\cite{skeleton_hand} to define a set of local coordinate systems with four level bones, and get the local transformation matrix $\mathbf{B}_{l}$ 
to map each 3D root-aligned joint to canonical joint. 
Details can be found in HALO~\cite{skeleton_hand}.
Following a similar approach, we can also calculate the corresponding pose parameter $\theta$ and bone length $\mathbf{L}$ with respect to hand skeleton pose $\mathbf{J}$.

During offline stage (Sec.~3.1 of our main paper), we first get bone length $\mathbf{L}$ for each hand according to the hand skeleton pose estimation. Then we fix the bone length $\mathbf{L}$ in offline stage and online stage (Sec.~3.2 of our main paper). 

\subsection{Definition of Pose Parameters}
\vspace{1mm}
\noindent \textbf{Definition of Hand Pose Parameters.} During hand pose optimization, we refine the hand pose parameters $\theta_{hand}\in\mathbb{R}^{36}$ including the rotation and translation of the palm root joint $\mathbf{J}_{0}$, the angles between the normal directions of two adjacent palm planes (3 in total), the angles between adjacent level-0 bones (4 in total), the flexion angle on level 2-3 bones (10 in total, because the hand joints on level 2-3 bones have no degree of freedom in abduction angle.), and flexion and abduction angles on level 1 bones (10 in total).
We follow~\cite{continuity} to convert the rotation matrix $\mathbf{R}_{palm}\in\mathbb{R}^{3\times3}$ of the palm root to a 6D representation, and the translation of the root joint $\mathbf{T}_{palm}$ has three dimensions.
Therefore, a total of 36 hand pose parameters $\theta_{hand}\in\mathbb{R}^{36}$ will be optimized in the offline and online stages.

\vspace{1mm}
\noindent \textbf{Definition of Object Pose Parameters.} During the optimization of object pose, we optimize the rotation $\mathbf{R}_{o}\in\mathbf{R}^{3\times3}$, which is defined in 6D representation, and translation $\mathbf{T}_{o}\in\mathbb{R}^{3}$ of the object. Then a total of 9 object pose parameters $\theta_{object}\in\mathbb{R}^{9}$ will be optimized in the offline and online stages. 

\subsection{Pose Optimization in Offline and Online Stages}
During the offline stage, 
we have the estimated hand skeleton pose $\bar{\mathbf{J}}$ and object 6D pose $\bar{\mathbf{R}}_{o}, \bar{\mathbf{T}}_{o}$. We decouple $\bar{\mathbf{J}}$ into $\theta_{hand}$ and $\bar{\mathbf{L}}$ and use the forward kinematics function $H(\theta_{hand}, \mathbf{L})\mapsto(\mathbf{J},\mathbf{B}^{-1},\mathbf{T})$ to get bone transformation $\mathbf{B}^{-1}$ with hand pose parameter $\theta_{hand}$ and bone length $\mathbf{L}$ (Sec.~3.1 of the main paper), and the bone length $\mathbf{L}$ of each hand is fixed during optimization.
In order to reduce the inevitable pose errors in the offline stage, we obtain the final hand pose  $\mathbf{J}$ and object pose $\mathbf{R}_{o}, \mathbf{T}_{o}$ by optimizing the loss function in the offline stage (Eq.~5 in the main paper) with respect to $\theta_{hand}$, $\theta_{object}$ using $\Delta\theta_{hand}$, $\Delta\mathbf{R}_{o}$, $\Delta\mathbf{T}_{o}$, and the other network parameters of hand and object models.

In order to conduct joint model fitting at online stage, we adopt images of sparse camera views at each frame for fitting.
We first use LT~\cite{learnable-triangulation} to get initial hand pose $\hat{\mathbf{J}}$, use Cosypose~\cite{cosypose} to get initial object 6D pose $\hat{\mathbf{R}}_{o}, \hat{\mathbf{T}}_{o}$, and then fix the offline hand and object models and get
the final hand pose  $\mathbf{J}$ and object pose $\mathbf{R}_{o}, \mathbf{T}_{o}$ by optimizing the online stage loss (Eq.~7 or Eq.~8 in the main paper) with respect to $\theta_{hand}$ and  $\theta_{object}$, respectively.
Since our offline object model can get object mesh model $\mathbf{V}_o$ in the object coordinate system, we fix the mesh model of each object before fitting. 
Then the object vertices under object pose $\mathbf{R}_{o}, \mathbf{T}_{o}$ can be calculated by $\mathbf{V}=\mathbf{R}_{o}\mathbf{V}_o+\mathbf{T}_{o}$, and they can be used to calculate the stable contact loss in Sec.~3.2.3 of our main paper. 

\subsection{Details of Model Fitting for Video}
In Sec.~3.2.3 of our main paper, we present loss function of joint model fitting for video sequence.  
However, fitting on a video sequence is often stuck in local minima due to its non-convex property with high dimension~\cite{hampali2020honnotate}, which can lead to noisy updates of $\theta_{hand}$ and $\theta_{object}$~\cite{a-nerf}. We adopt a divided-and-conquer optimization strategy based on sliding window.
In each iteration, we select a sliding window with four adjacent frames, and optimize the pose of each frame in the sliding window. The frames in the window are fitted, and continued to be optimized for four times.
Then the optimization switches to the next time window.
After the optimization of the time window at the end of the video is conducted, we return to the first time window to start a new optimization iteration using sliding window. In the experiment, the optimization will iterate over the entire sequence for 5 times. 

\section{Additional Experimental Results}
\label{sec:supp_2}

\begin{figure*}[th]
\centering%
\includegraphics[width=\linewidth]{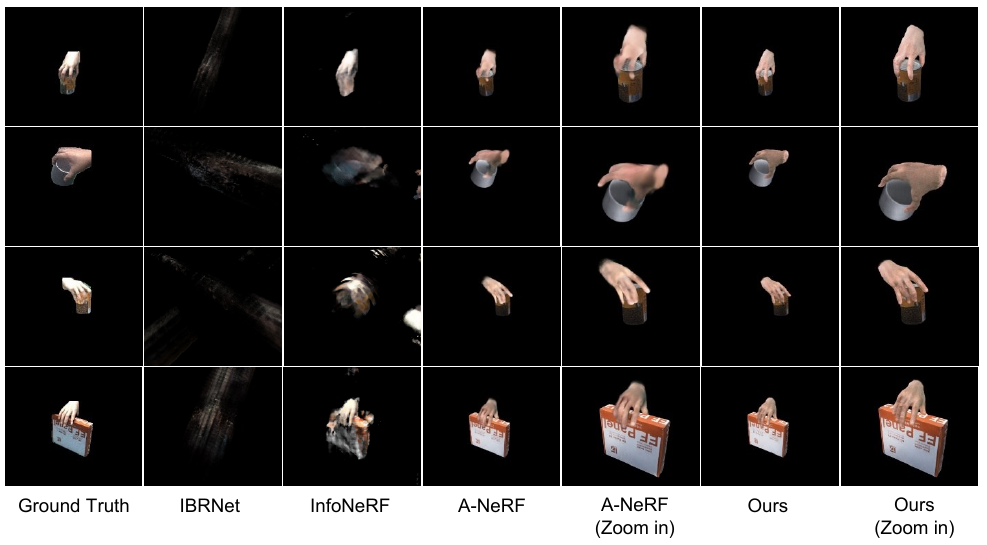}
\caption{Rendering quality comparison on the HandObject dataset under three camera views with SoTA methods. Our pre-built models preserve the shape and appearance priors and achieves better rendering results from sparse views. We zoom in the rendering results for demonstration.}
\label{fig:supp_render_quality_sota}
\end{figure*}
\subsection{Comparison Results}
\vspace{1mm}
\noindent \textbf{Comparison on Rendering Quality with SoTA Methods.} Fig.~\ref{fig:supp_render_quality_sota} shows more qualitative comparison with the SoTA methods, A-NeRF~\cite{a-nerf}, IBRNet~\cite{ibrnet} and InfoNeRF~\cite{infonerf} on HandObject dataset under 5 test views. We find that few-shot neural rendering methods such as IBRNet~\cite{ibrnet} and InfoNeRF~\cite{infonerf} cannot work well when the camera views are widely separated. Our two-stage method can achieve better results because the pre-built hand and object models preserve the shape and appearance priors. We replace our hand model with A-NeRF based hand model and use the same object model for fitting and rendering. Compared to the density representation in A-NeRF, the SDF representation in our model can get high-quality appearance,
and the rendering quality of our method is better (See Table 4 of our main paper).
\begin{figure}[th]
\centering%
\includegraphics[width=\linewidth]{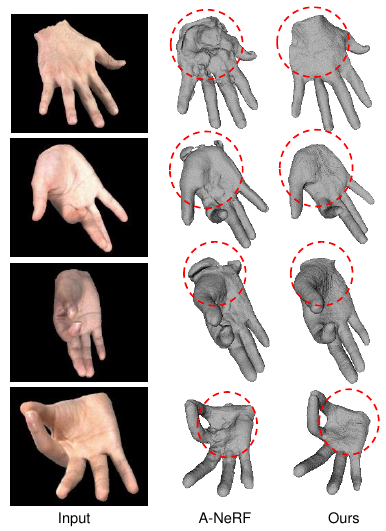}
\caption{Hand surface comparison with A-NeRF~\cite{a-nerf}. Our method can achieve much better hand surface reconstruction results, and the red circle highlights that our method has less shape artifacts.}
\label{fig:supp_reconstruction}
\end{figure}
\begin{figure}[h]
\centering%
\includegraphics[width=\linewidth]{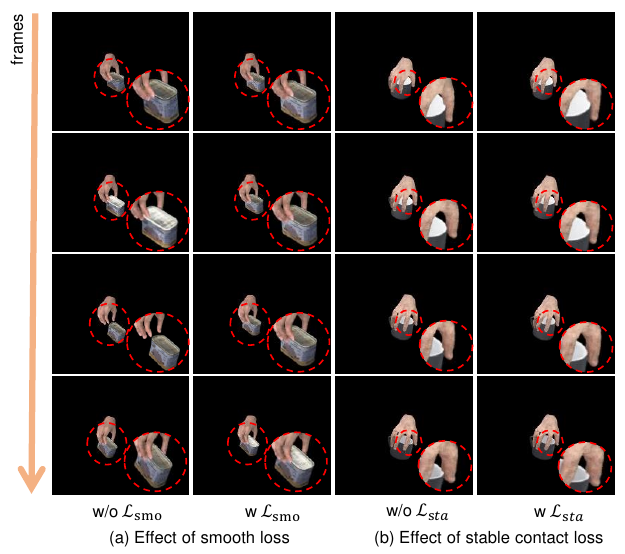}
\caption{Effect of smooth and stable contact loss. (a) The jitters are significantly reduced by adding smooth loss. The jitter at the object is significantly reduced when smooth loss is added. (b) The contact regions are more stable with stable contact loss.
After adding stable contact loss, the contact regions are enforced to be more stable, and sliding effects are effectively reduced.
}
\label{fig:supp_stable_smooth}
\end{figure}

\begin{figure*}[h]
\centering%
\includegraphics[width=\linewidth]{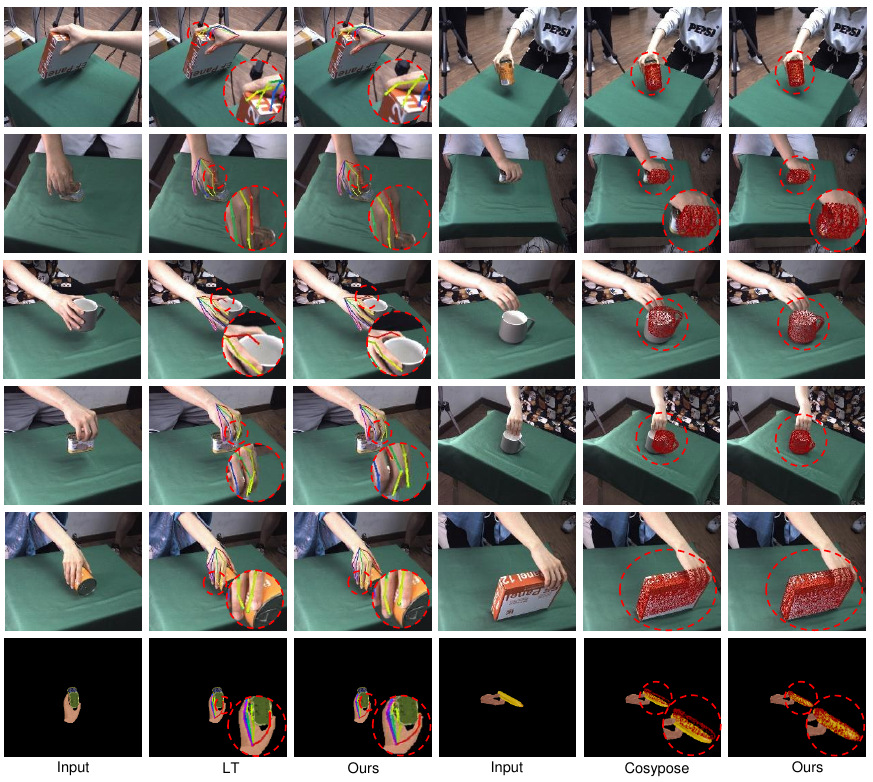}
\caption{Effect of model fitting on pose estimation. After fitting, the projected hand skeleton pose is aligned well with the hand, and the projected object mesh model are also well aligned with the object boundary.
}
\label{fig:supp_pose_refine}
\end{figure*}

\vspace{1mm} 
\noindent \textbf{Effect of Model Fitting on Pose Estimation.} In order to investigate whether model fitting is effective to improve pose estimation during online stage, we compare pose performance with initial hand pose by LT~\cite{learnable-triangulation}, GHPT~\cite{ghpt}, I2L~\cite{i2l} and initial object pose by Cosypose~\cite{cosypose} as shown in Table 1 of our main paper.
We observe that better hand skeleton pose and object pose can be achieved with our online model fitting (i.e. 'GHPT+CP+Ours', 'LT+CP+Ours'). 
Fig.~\ref{fig:supp_pose_refine} shows qualitative comparison
on pose estimation and the refined pose with model fitting under 8 camera views in HandObject dataset compared with LT and CosyPose. After fitting the projected hand skeleton pose is well aligned with the hand, and the projected object mesh model are also well aligned with the object boundary. 

\begin{table}[ht]
\centering
\resizebox{0.8\linewidth}{!}{
\begin{tabular}{c|c|c|c}
\hline
Method & PSNR  $\uparrow$& SSIM  $\uparrow$& LPIPS $\downarrow$\\\cline{1-4}
A-NeRF~\cite{a-nerf} & 18.61 & 76.51 & 0.240 \\ 
Ours& \textbf{19.85} & \textbf{79.66} & \textbf{0.158}\\

\hline
\end{tabular}
}
\caption{Quantitative comparison with A-NeRF on the rendering quality of hand models in HandObject dataset. Our hand model outperforms A-NeRF based hand model.}
\label{table:supp_anerf_hand_render_quality}
\end{table}

\begin{figure}[th]
\centering%
\includegraphics[width=\linewidth]{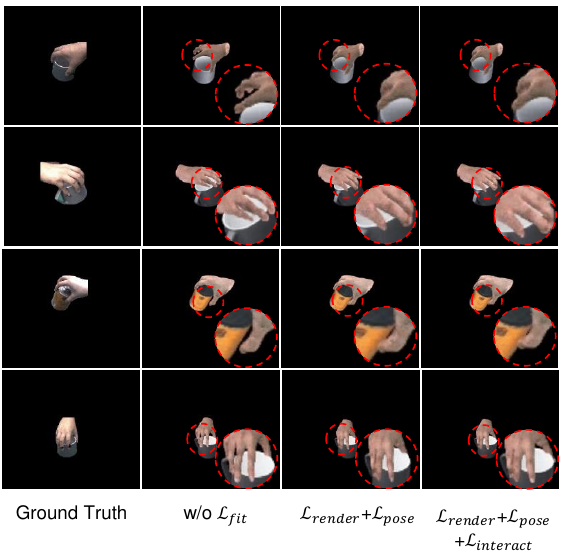}
\caption{Quantitative comparison of interaction loss on rendering quality. The accurate pose after fitting improves the rendering quality, and $\mathcal{L}_{interact}$ further improves the rendering quality by solving unreasonable interactions such as penetration.}
\label{fig:supp_render_quality_loss}
\end{figure}

\vspace{1mm} 
\noindent \textbf{Comparison with A-NeRF~\cite{a-nerf}.}
A-NeRF~\cite{a-nerf} is a generative neural body model, and we apply it to build the neural hand model, but it cannot be directly used to represent the hand-object interaction scene. We combine the A-NeRF based hand model with our object model to represent hand-object interaction scene.

\begin{figure*}[htb]
\centering%
\includegraphics[width=0.93\linewidth]{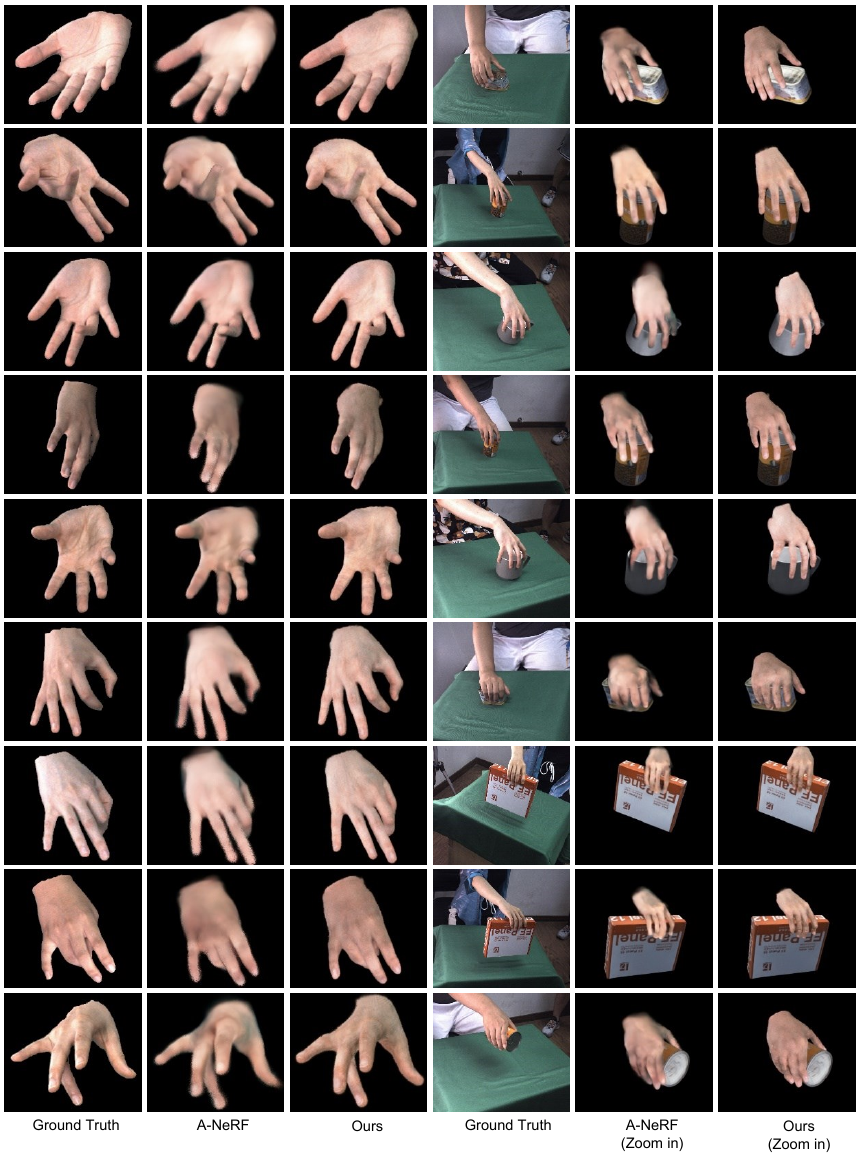}
\caption{Qualitative comparisons with the A-NeRF~\cite{a-nerf} based hand model. Our results preserve more texture details on the hand.
}
\label{fig:supp_com_anerf}
\end{figure*}

We first show quantitative comparison with A-NeRF on the rendering quality of hand models in HandObject dataset in Table~\ref{table:supp_anerf_hand_render_quality}, and our hand model outperforms A-NeRF based hand model on rendering quality evaluation metrics.
Then we replace our hand model with A-NeRF based hand model in hand-object interaction scene. Table 4 of our main paper shows that our model can achieve better rendering quality than A-NeRF based hand model in hand-object interaction. 
Fig.~\ref{fig:supp_com_anerf} shows more qualitative comparisons on hand and hand-object interaction with A-NeRF based hand model. We observe that our results preserve more appearance details on the hand.

In order to investigate the effect of rendering quality on pose optimization, we compare pose accuracy after joint model fitting with our method and A-NeRF based pre-built hand model in HandObject dataset under 8 camera views.  
As shown in Table 1 of our main paper, high-quality rendering results of the hand model with our method are conducive to obtaining more accurate poses with the rendering-based optimization than A-NeRF based method (i.e. 'LT+CP+A-NeRF'). 

Fig.~\ref{fig:supp_reconstruction} shows the qualitative comparison on hand reconstruction with our method and A-NeRF. Our method can achieve better hand surface reconstruction results, and the red circle highlights that our method has less shape artifacts.

\vspace{1mm}
\noindent \textbf{Comparison with the Parametric Model on Pose Optimization.}
We conduct comparison experiments using parametric models for pose optimization. We first use I2L~\cite{i2l} to estimate the pose and shape parameters for hand parametric model MANO~\cite{mano} and use MANO hand mesh and the object mesh obtained in the offline stage for pose optimization. We use images in HandObject dataset under 8 camera views for fitting. We do not use the color loss for pose optimization with MANO due to the lack of texture in MANO. Table 1 of our main paper shows that the parametric model ﬁtting cannot achieve accurate pose (i.e. 'I2L+CP+Mesh Fitting'). Conceptually, the mask loss is the main loss for pose optimization with MANO, and it is inferior to the color loss to provide sufficient constraints to achieve accurate pose results. 

\noindent \textbf{Pose Estimation Baseline}. In hand pose estimation, we compare with the state-of-the-art (SoTA) multi-view pose estimation method including GHPT~\cite{ghpt} and LT~\cite{learnable-triangulation} and a single view pose estimation method  I2L~\cite{i2l}. We use MMPose~\cite{mmpose} to obtain the 2D initial hand pose for GHPT. I2L is used to predict the parameters of the hand model MANO~\cite{mano} in a single view, and we extend it to multi-view task. We transform the MANO parameters obtained by multi-view images from camera coordinate to world coordinate and average the MANO parameters to obtain the final parameters in the world coordinate system, and then obtain the translation of the wrist in the world coordinate system based on the triangulation of the predicted 2D key points of the wrist.
In object pose estimation, we compare with the SoTA multi-view method CosyPose~\cite{cosypose} and initialize the pose from the output of PoseCNN~\cite{posecnn} for CosyPose.

\noindent \textbf{Dataset Details in Pose Estimation Comparison.} In HandObject dataset, we choose 50 interaction sequences as the test set. We select 4 single-hand data and 12 interaction sequences as training dataset for hand pose estimation. We select 4 single-object data and 25 interaction sequences as training dataset for object pose estimation. 
In Synthetic DexYCB dataset, we choose 20 interaction sequences as the test set. We select 4 single-hand data and 36 interaction sequences as training dataset for hand pose estimation. We select 5 single-object data and 36 interaction sequences as training dataset for object pose estimation.

\noindent \textbf{Novel View Synthesis Baseline}. We select frames from the test set in HandObject dataset, of which 3 views are used for fitting or training few-shot methods, and the remaining 5 views are used for testing. For training IBRNet~\cite{ibrnet}, We use the official parameter model and reﬁne it for each input frame. For training InfoNeRF~\cite{infonerf}, We train a model from scratch for each input frame.

\noindent \textbf{A-NeRF~\cite{a-nerf} Baseline}. We select single-hand images to train the A-NeRF based hand models for four subjects, which are the same as used in our offline hand models. 

\subsection{Ablation Study}

\vspace{1mm}
\noindent \textbf{Effect of Smooth Loss and Stable Contact Loss.}
We show qualitative results on smooth loss and stable contact loss in HandObject dataset under 8 camera views in Fig.~\ref{fig:supp_stable_smooth}.
We observe that the jitter at the object is significantly reduced when smooth loss is added (Fig.~\ref{fig:supp_stable_smooth}(a)).
As shown in Fig.~\ref{fig:supp_stable_smooth}(b), the hand object contact is more consistent with stable contact loss.
Due to heavy inter-occlusions between hand and object, it is easy to get results with fingers sliding on connecting surfaces.
Since the stable contact loss integrates the contact area information between frames, especially at the edges of object, after adding stable contact loss, the contact regions are enforced to be more stable, and sliding effects are effectively mitigated. More results can be found in the video.

\begin{figure*}[h]
\centering%
\includegraphics[width=\linewidth]{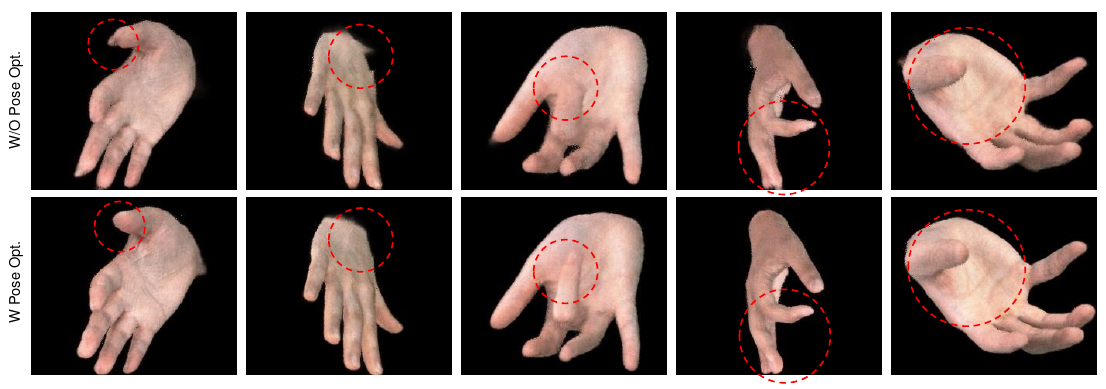}
\caption{Effect of pose optimization on hand model in offline stage. 
After pose optimization, the hand model learns more realistic texture details, and the rendering results are especially better around finger joints.
}
\label{fig:supp_offline_pose_opt}
\end{figure*}

\vspace{1mm}
\noindent \textbf{Effect of Pose Optimization on Hand Model in Offline Stage.} In order to investigate the effect of pose optimization on hand model in offline stage, we compare hand rendering results without pose optimization.
Fig.~\ref{fig:supp_offline_pose_opt} shows the qualitative results of pose optimization. We observe that through pose optimization, the rendering results of the hand model are more realistic.
More results can be found in the video.

\vspace{1mm}
\noindent \textbf{Effect of Interaction Loss on Rendering Quality.}
We show the qualitative comparison results on rendering quality under different loss combinations in Table 7 of our main paper. We show the results of the quantitative comparison in Fig.~\ref{fig:supp_render_quality_loss}.
The red circle shows that after fitting (i.e. '$\mathcal{L}_{render}+ \mathcal{L}_{pose}$', '$\mathcal{L}_{render}+\mathcal{L}_{pose} +\mathcal{L}_{interact}$'), the pose results of hands and objects become more accurate (referring to Table 2 and Table 3 of our main paper), and the corresponding rendering quality evaluation metrics are improved (referring to Table 7 of our main paper).
The interaction loss $\mathcal{L}_{interact}$ can further improve the rendering quality, because the incorrect color caused by unreasonable interactions such as penetration can be reduced.

\subsection{Scene Editing Results}
\begin{figure*}[h]
\centering%
\includegraphics[width=\linewidth]{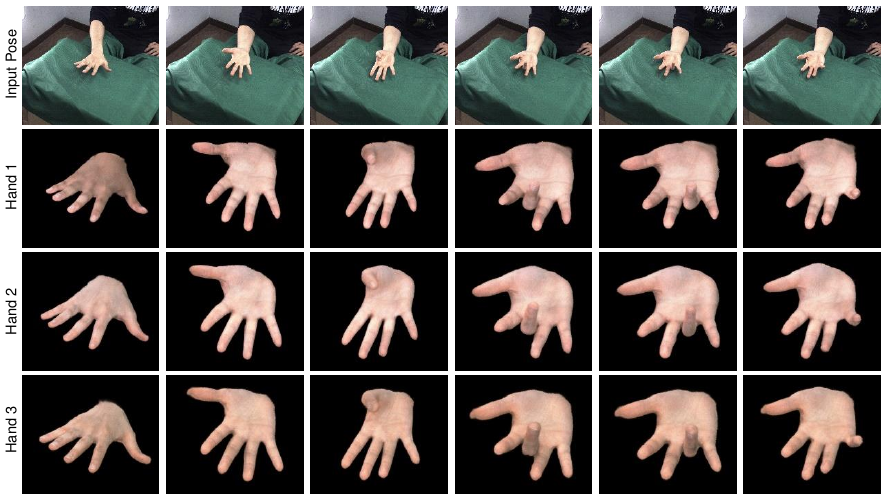}
\caption{Our pose editing results. We can get new rendering results using our hand model by changing the hand skeleton pose and we show two pose-driven hand models under various poses.}
\label{fig:supp_pose_editing}
\end{figure*}

\vspace{1mm}
\noindent \textbf{Editing Hand Pose.} We can get new rendering results using our hand model by changing the hand skeleton pose. Our hand model can be driven by various poses and the results are shown in Fig.~\ref{fig:supp_pose_editing}. We use the same pose sequence to edit on three different hand models.
More results can be found in the video.

\begin{figure*}[h]
\centering%
\includegraphics[width=0.92\linewidth]{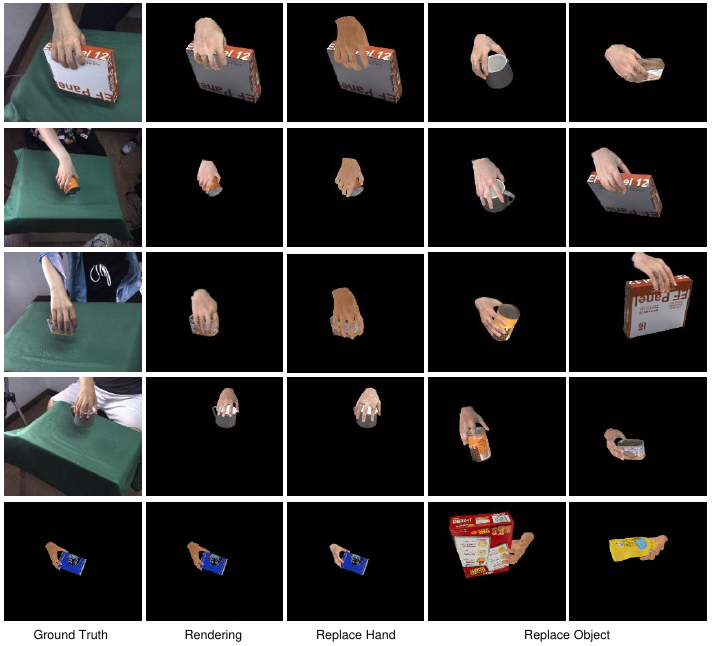}
\caption{Rendering results by replacing the hand and object models.
In the third column, we replace the hand model and get realistic rendering results. In the last two columns, we change the object model and the pose of hand to edit the entire scene.
}
\label{fig:supp_model_edting}
\end{figure*}

\noindent \textbf{Replacing Models.} We can replace the hand or object model and get the corresponding results (Fig.~\ref{fig:supp_model_edting}). In the third column of Fig.~\ref{fig:supp_model_edting}, we replace the hand model and get realistic rendering results. We can also change the object model and the pose of hand to edit the interaction scene.

\subsection{Novel View Synthesis and Reconstruction}
\begin{figure*}[htp]
\centering%
\includegraphics[width=\linewidth]{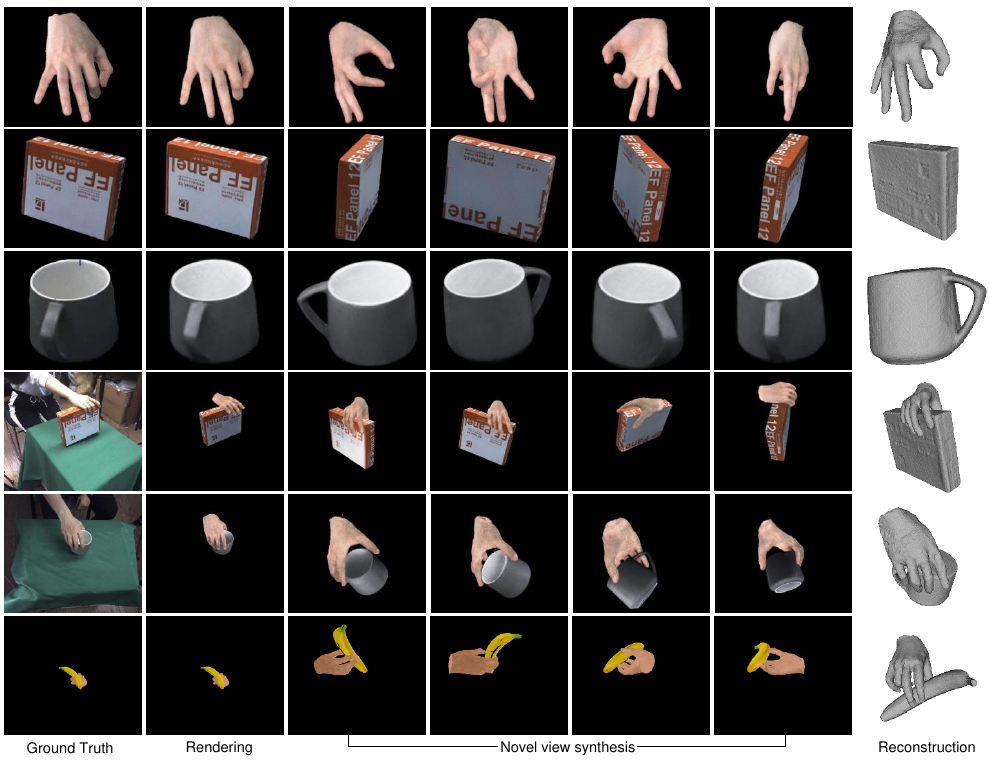}
\caption{The results on novel view synthesis and reconstruction. The rendering results of our hand and object models preserve realistic texture information and enable full 360 degree free-viewpoint rendering, and we can get high-quality hand-object interaction reconstruction results.}
\label{fig:supp_novel_view_re}
\end{figure*}

Fig.~\ref{fig:supp_novel_view_re} shows more novel view synthesis and reconstruction results at the offline stage and the online stage.
The rendering results with our hand and object models preserve realistic details and enable full 360 degree free-viewpoint rendering and we can get high-quality hand-object interaction reconstruction results.

\section{Limitation and Failure Case}
\label{sec:supp_3}
\begin{figure}[th]
\centering%
\includegraphics[width=\linewidth]{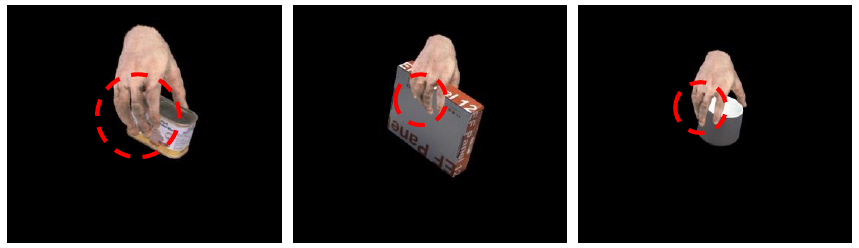}
\caption{Failure case of our method. Our model does not take into account the influence of shadow on the hand, so that the rendering results in random perspectives may contain unrealistic shadows.}
\label{fig:supp_failure_case}
\end{figure}

Although promising results can be achieved with our proposed method, there are several key challenges to be solved in the future study. First, our model does not take into account the influence of shadow on the hand, so that the rendering results in random perspectives may contain unrealistic shadows. We show the failure cases in Fig.~\ref{fig:supp_failure_case}. Second, in the process of hand pose optimization, the self-penetration problem of the hand is not considered, which gets the self-penetration between fingers occasionally. Finally, the rendering efficiency of the model should be improved in the future. 

\clearpage


\end{document}